\documentclass[preprint,5p,times,twocolumn]{elsarticle}

\usepackage{amssymb}
\usepackage{amsmath}
\usepackage{multirow}

\usepackage{booktabs} 
\usepackage{xcolor, colortbl}
\usepackage{capt-of}
\usepackage{amssymb}
\usepackage{amsfonts}
\definecolor{cgreen}{rgb}{0,0.7,0.8}
\definecolor{cred}{rgb}{0.968,0.545,0.321}

\newcommand{\rtr}[1]{{\scriptsize\color{cred}$\blacktriangledown$ #1}}

\newcommand{\gtr}[1]{{\scriptsize\color{cgreen}$\blacktriangle$ #1}}
\usepackage{float}
\usepackage{svg}
\usepackage{booktabs}   
\usepackage{caption}    
\usepackage{siunitx}    
\usepackage{makecell}   
\usepackage{url}

\journal{Medical Image Analysis}

\begin{document}

\begin{frontmatter}



\title{Progressive Growing of Patch Size: Curriculum Learning for Accelerated and Improved Medical Image Segmentation}


\author[aff1,aff2,aff3,aff4]{Stefan M. Fischer\fnref{clabel}}
\ead{stefan.mi.fischer@tum.de}
\author[aff1,aff2,aff3,aff4]{Johannes Kiechle}

\author[aff1,aff3]{Laura Daza}
\author[aff1,aff3]{Lina Felsner}
\author[aff1,aff3,aff6]{Richard Osuala}
\author[aff1,aff3]{Daniel M. Lang}
\author[aff6]{Karim~Lekadir}

\author[aff3, aff7]{Jan C. Peeken \fnref{slabel}}
\author[aff1,aff3,aff4,aff5]{Julia A. Schnabel \fnref{slabel}}
\fntext[clabel]{Stefan M. Fischer is corresponding author}
\fntext[slabel]{Jan C. Peeken and Julia A. Schnabel contributed equally as senior authors.}

\affiliation[aff1]{organization={School of Computation, Information and Technology, Technical University Munich},
city={Munich},
country={Germany}}

\affiliation[aff2]{organization={Department of Radiation Oncology, TUM School of Medicine, TUM University Hospital rechts der Isar, Technical University of Munich},
city={Munich},
country={Germany}}

\affiliation[aff3]{organization={Institute of Machine Learning in Biomedical Imaging, Helmholtz Munich},
city={Munich},
country={Germany}}

\affiliation[aff7]{organization={Institute of Radiation Medicine, Helmholtz Center Munich},
            city={Munich},
            country={Germany}}

\affiliation[aff4]{organization={Munich Center of Machine Learning (MCML)},
city={Munich},
country={Germany}}

\affiliation[aff5]{organization={School of Biomedical Engineering and Imaging Sciences, King’s College London},
city={London},
country={UK}}

\affiliation[aff6]{organization={Departament de Matemàtiques i Informàtica, Barcelona Artificial Intelligence in Medicine Lab (BCN-AIM), Universitat de Barcelona},
city={Barcelona},
country={Spain}}

\begin{abstract}
In this work, we introduce Progressive Growing of Patch Size, an automatic curriculum learning approach for 3D medical image segmentation. Our approach progressively increases the patch size during model training, resulting in an improved class balance for smaller patch sizes and accelerated convergence of the training process. We evaluate our curriculum approach in two settings: a resource-efficient mode and a performance mode, both regarding segmentation performance with respect to the Dice score and computational costs across 15 diverse and popular 3D medical image segmentation tasks. The resource-efficient mode matches the segmentation performance of the conventional constant patch size sampling baseline with a notable reduction in training time to only 44\%. The performance mode improves upon constant patch size segmentation results, achieving a statistically significant relative mean performance gain of 1.28\% in Dice score. Remarkably, across all 15 tasks, our proposed performance mode manages to surpass the constant patch size baseline in segmentation performance, while simultaneously reducing training time to only 89\%. The benefits are particularly pronounced for highly imbalanced tasks such as lesion segmentation tasks. Rigorous experiments demonstrate that our performance mode not only improves mean segmentation performance but also reduces performance variance, yielding more trustworthy model comparison. Furthermore, our findings reveal that the proposed curriculum sampling is not tied to a specific architecture but represents a broadly applicable strategy that consistently boosts performance across diverse segmentation models, including UNet, UNETR, and SwinUNETR. In summary, we show that this simple yet elegant transformation on input data substantially improves both segmentation performance and training runtime, while being compatible across diverse segmentation backbones. 

\end{abstract}

\begin{graphicalabstract}
\centering
\includegraphics[width=0.9\textwidth]{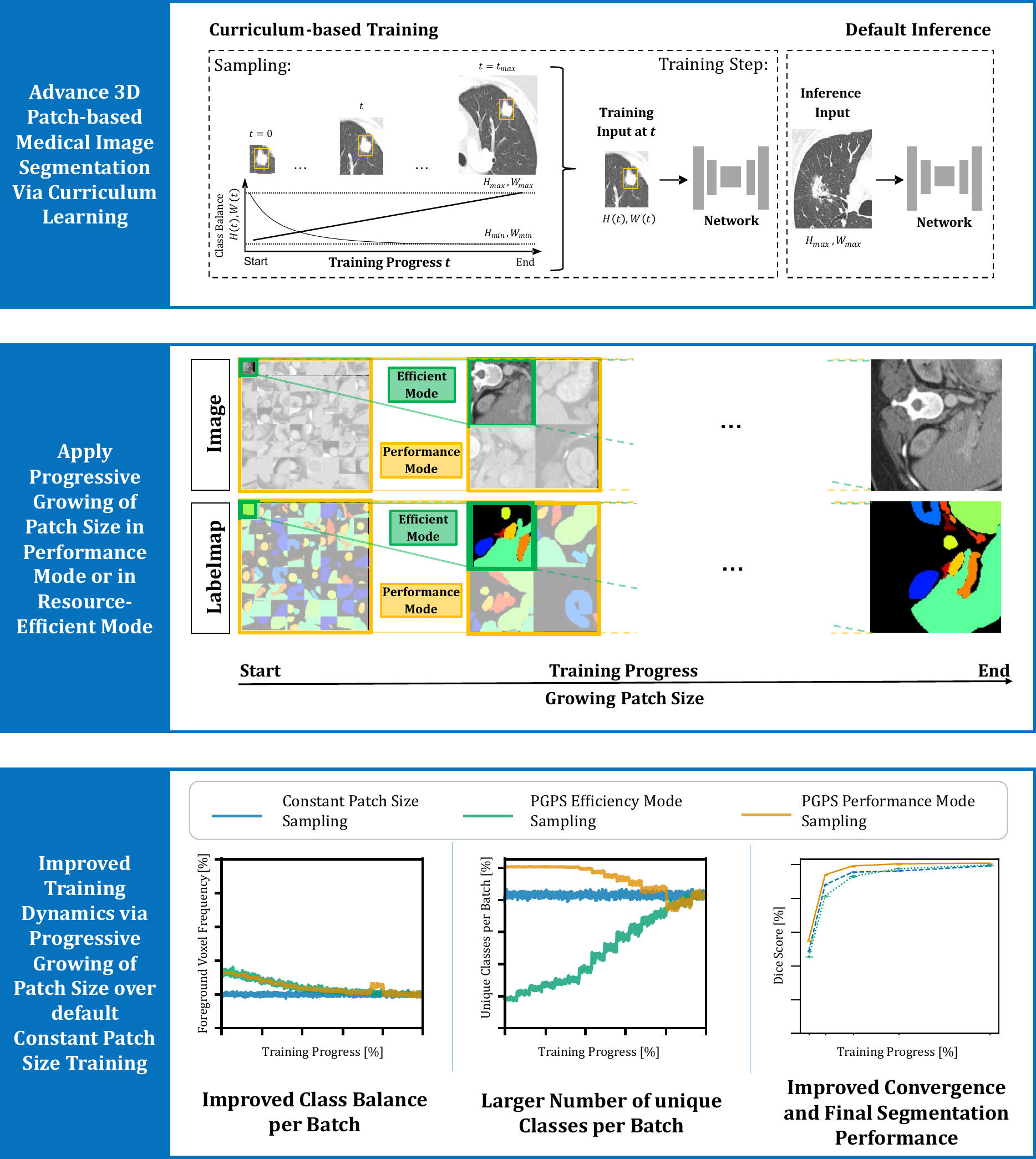}
\end{graphicalabstract}

\begin{highlights}
\item Novel patch-size curriculum improves performance over fixed patch size training. 
\item Performance mode boosts Dice scores across all datasets with lower training cost.  
\item Efficiency mode matches Dice scores while cutting runtime and FLOPs sharply.  
\item Patch-size curriculum implicitly improves class balance for segmentation tasks. 
\item  Simple to apply within standard patch-based 3D medical segmentation setup.

\end{highlights}


\end{frontmatter}




\section{Introduction}

Most research efforts in the area of medical image segmentation focus on the development of new architectural concepts, including convolution-based~\cite{ronneberger2015u,isensee2021nnu}, transformer-based \cite{dosovitskiy2020image,liu2021swin}, and hybrid approaches \cite{hatamizadeh2022unetr,hatamizadeh2021swin}. In contrast, comparatively little attention has been given to the training process itself, where models are still predominantly trained using random data sampling strategies. Inspired by human learning, Bengio \textit{et al.}~\cite{bengio2009curriculum} have introduced the concept of curriculum learning, where models first learn simpler tasks before progressing to more difficult ones. By ordering samples from easy to hard, Bengio \textit{et al.}~\cite{bengio2009curriculum}have demonstrated faster convergence and improved performance relative to random sampling. Such techniques also have the potential to reduce the substantial computational costs of training, both in terms of time and energy consumption, leading to a beneficial environmental impact~\cite{selvan2022carbon}.

Several approaches have been proposed to design curricula for medical imaging. Some rely on human annotations or expert knowledge to define task-specific measures of difficulty~\cite{bengio2009curriculum,jimenez2019medical,wei2021learn}. For example, task difficulty might be linked to class membership, as in fracture classification~\cite{jimenez2019medical}, or defined using inter-rater agreement~\cite{jimenez2019medical,wei2021learn}. However, expert annotations considerably increase the costs, making those approaches often infeasible. More general strategies estimate task difficulty automatically or synthetically change data complexity during training. Sample loss has been used as a proxy for difficulty, enabling hard-negative mining and adaptive oversampling, as shown in lung cancer segmentation~\cite{jesson2017cased}. In MRI-based brain tumor segmentation, Havaei~\textit{et al.}~\cite{havaei2016hemis} have introduced a curriculum that improves robustness to missing modalities by randomly dropping channels with increasing probability during training.  
%
Karras~\textit{et al.}~\cite{karras2017progressive} have proposed to progressively grow a generative adversarial network layer by layer, effectively increasing task difficulty as output resolution doubled. This approach has been used for natural image generation and has shown improved convergence, reduced training time, and an overall better model performance. Zhao~\textit{et al.}~\cite{zhao2020pgu} have extended this idea to semantic segmentation of cervical nuclei by progressively growing the UNet architecture from its bottleneck.  

Another way to define task difficulty is by sample length. In natural language processing, curricula based on sentence length have been used to accelerate the training of large language models such as BERT and GPT-2~\cite{devlin2019bert, li2022stability, press2020shortformer, shen2024efficient} and have been adopted into the high-performance training library DeepSpeed~\cite{li2024deepspeed}. In computer vision, an analogous measure is image size. Several works have employed curricula that (progressively) increase input resolution~\cite{tan2021efficientnetv2, wang2024efficienttrain++, bolya2025perceptionencoderbestvisual, shen2024efficient} during training. These methods consistently improve convergence, enabling either more resource-efficient training with comparable performance~\cite{tan2021efficientnetv2, wang2024efficienttrain++} or higher final accuracy within the same training budget~\cite{bolya2025perceptionencoderbestvisual}. Several curricula have been applied for visual backbone pretraining tasks~\cite{bolya2025perceptionencoderbestvisual, wang2024efficienttrain++}, by pretraining on ImageNet classification~\cite{wang2024efficienttrain++, tan2021efficientnetv2} or contrastive pretraining~\cite{bolya2025perceptionencoderbestvisual, lireclip, koccyiugit2023accelerating}. In contrast, there is limited work in applying the sample-length curriculum directly to segmentation tasks~\cite{arani2021rgpnet}, thus semantic segmentation mostly benefits from curriculum learning only indirectly via transferring pretrained weights~\cite{bolya2025perceptionencoderbestvisual, wang2024efficienttrain++}.

In this work, we propose a Progressive Growing of Patch Size curriculum, which is a novel sample-length curriculum for 3D medical image segmentation, defined through patch size. Our approach starts training on small patches, ensuring strong class balance in early training, and progressively increases to large patches, which provide a broader global context. Furthermore, the reduced memory footprint of smaller patches early in training allows for larger batch sizes under the same GPU budget. We argue that our proposed curriculum, built on patch size, is better suited for dense prediction tasks such as semantic segmentation than the Progressive Resolution curriculum applied in multiple methods in the CV domain~\cite{tan2021efficientnetv2, wang2024efficienttrain++, bolya2025perceptionencoderbestvisual}. Furthermore, our approach is an advancement over the conventionally applied constant patch size sampling.

We have introduced the key concept of Progressive Growing of Patch Size (PGPS) in our previous work at MICCAI 2024~\cite{fischer2024progressive}, where we have established the fundamental idea of progressively increasing patch size during training. In this work, we significantly advance the initial concept by optimizing the methodology for computational efficiency and segmentation performance, and largely broadening the experimental and analytical scope in the following ways: (i) We propose a performance mode of the curriculum resulting in improved segmentation performance beyond the originally proposed method in \cite{fischer2024progressive}; (ii) We have investigated the relationship between dataset characteristics and the observed performance gains, identifying improved class balance as the key factor driving the enhanced results.; (iii) We have proven the generalization to different backbones by applying it successfully to one fully convolutional network, UNet, and two transformer-based networks, UNETR and SwinUNETR; (iv) We have analyzed the impact of stochastic training variability and show that the proposed performance mode notably reduces segmentation outcome variance compared to conventional constant patch size training; (v) We have conducted a direct comparison between our patch size–based curriculum and an existing resolution-based curriculum, showing that our approach consistently achieves superior segmentation performance..
In this work, we show that our proposed curriculum, built on patch size, improves segmentation performance in the form of Dice score, and at the same time reduces the computational costs of training compared to conventional constant patch size training.

\begin{figure*}[th!]
\centering
\includegraphics[width=0.9\textwidth]{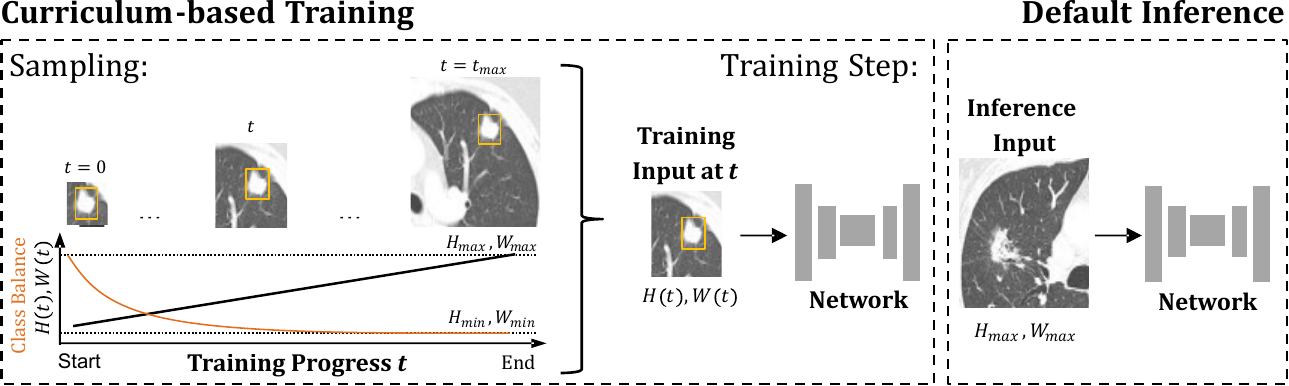}
\caption{Overview of the proposed \textbf{Progressive Growing of Patch Size} curriculum, illustrated for lung cancer segmentation (cancer regions highlighted with yellow bounding boxes). Training begins with the minimal patch size ($H_{min}, W_{min}$) and progressively increases the patch dimensions stepwise until the final maximal patch size ($H_{max}, W_{max}$) is reached. Smaller patch sizes provide a better class balance, which decreases as the patch size grows. During inference, the maximum patch size is used to capture maximal global context. Figure is adapted from~\cite{fischer2024progressive}.}
\label{fig:curriculum-sketch}
\end{figure*}

\section{Methods}

This section describes the proposed methods and is split into four parts. As a basis, we describe conventional ways of sampling in patch-based medical segmentation. Then, we discuss the theoretical design of the proposed curriculum learning. After that, we introduce the efficient curriculum mode and the performance mode. Finally, we outline the nnU-Net framework implementation of the curricula. 

\subsection{Sampling in Patch-based 3D Medical Image Segmentation}
\label{sec:background}

Three-dimensional (3D) medical image segmentation is fundamentally constrained by the large GPU memory requirements of volumetric data processing. Common strategies to address this limitation include: (1) using low-resolution models~\cite{isensee2021nnu}, (2) applying high-resolution patch-based methods~\cite{hatamizadeh2021swin, hatamizadeh2022unetr, roy2023mednext, isensee2021nnu, ma2024u}, and (3) employing network cascades~\cite{isensee2019automated} or multi-scale approaches~\cite{KAMNITSAS201761}. Among these, it has been shown that high-resolution patch-based methods generally yield superior performance~\cite{isensee2019automated, jeon2025no}. Patch sampling strategies play a central role in patch-based methods. nnU-Net adopts a ``forced oversampling'' technique, where one foreground (FG) patch is sampled from a randomly chosen patient, with FG classes selected with equal probability, while another random patch is sampled from a different patient~\cite{isensee2021nnu}. In contrast, MONAI implements ``Probabilistic Oversampling'' which applies probabilistic class sampling~\cite{cardoso2022monai}. To maximize global context, standard practice is to use the largest possible patch size that fits into GPU memory, typically with a batch size of two~\cite{isensee2021nnu}. Even with additional computational resources, nnU-Net prioritizes maximizing patch size over increasing batch size, as exemplified in its larger encoder variants such as nnU-Net ResEnc M/L/XL~\cite{isensee2024nnu}.

\subsection{Progressive Growing of Patch Size}

The central principle of our proposed PGPS curriculum is to begin the training with the smallest processable patch size and progressively increase to larger patches. The rationale for starting with smaller patches lies in task difficulty: smaller patches inherently yield a more balanced foreground-to-background ratio, whereas this balance diminishes as patch size increases. In the theoretical case of a single-voxel patch, a batch containing one foreground and one background patch would achieve perfect class balance, with half of the voxels belonging to the foreground and half to the background. As the patch size grows, however, the distribution converges toward the full volume statistics, which are typically dominated by background voxels.  

Patch size increments are applied in the smallest feasible steps, with each axis adjusted independently. By gradually increasing the patch size in minimal increments, transitions between training stages are smoothed, as the segmentation tasks at successive patch sizes remain closely related. This design provides the network with a structured sequence of progressively more challenging tasks. An overview of this curriculum is illustrated in Figure~\ref{fig:curriculum-sketch}. 

Both the minimal patch size and the patch size increment depend on the input size constraints of the underlying network architecture. The progression continues until the largest possible patch size is reached, which is typically constrained by the GPU memory capacity or set to the network's default patch size. During inference, the largest patch size is employed to maximize global context, which usually achieves the highest segmentation performance~\cite{isensee2021nnu}.

\subsection{Curriculum Modes: Efficiency vs. Performance}
\label{sec:curriculum-modes}
\begin{figure*}[th!]
\centering
\includegraphics[width=0.9\textwidth]{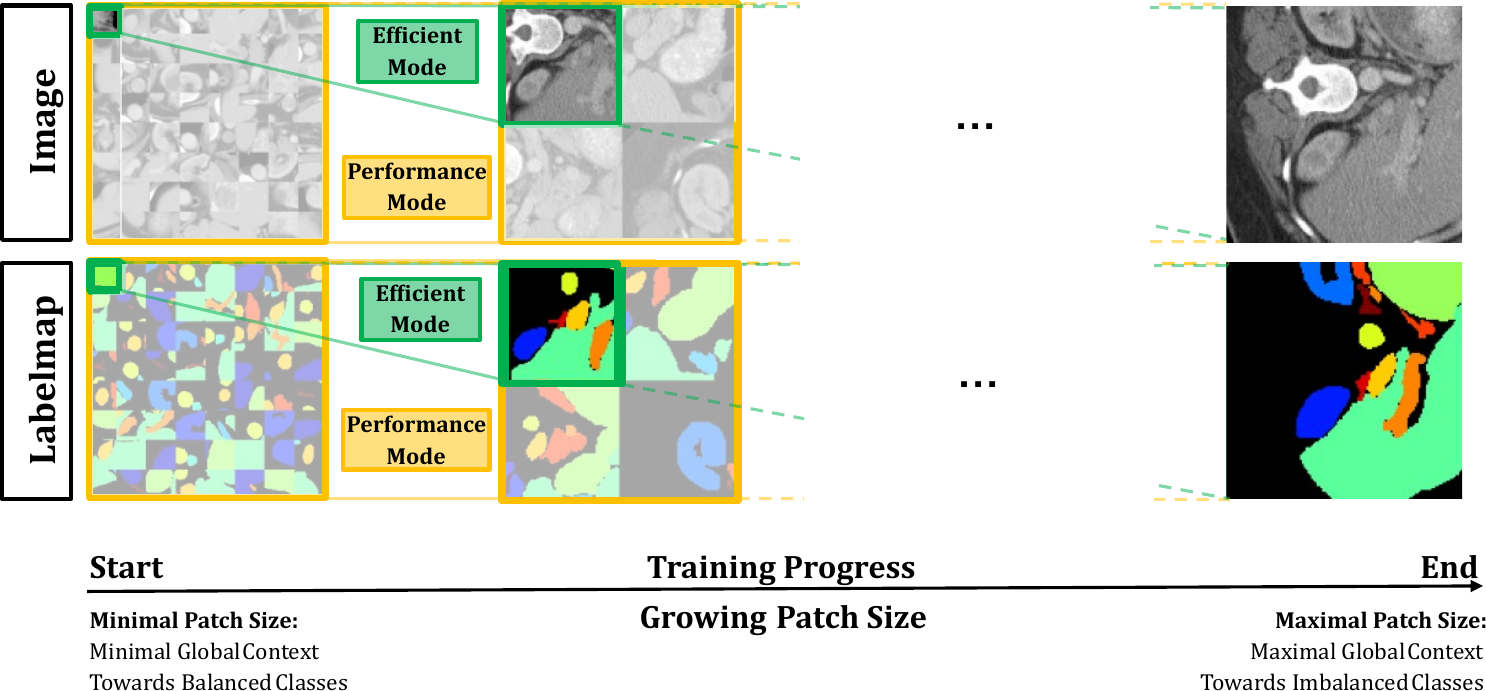}
\caption{Input tensors for the two proposed modes of the \textbf{Progressive Growing of Patch Size} (PGPS) curriculum, illustrated for a multi-organ segmentation task with example foreground patches (top) and corresponding label maps (bottom). In both modes, the patch size is progressively increased during training. In PGPS-Efficiency (green), the batch size remains constant, resulting on average in smaller input tensors. In PGPS-Performance (orange), the available GPU budget is fully utilized by increasing the batch size.}
\label{fig:perf-vs-eff}
\end{figure*}

We propose two modes of the PGPS curriculum, tailored to different objectives. \textbf{PGPS-Efficiency} minimizes training runtime while maintaining performance comparable to standard constant patch size sampling. \textbf{PGPS-Performance} aims to maximize segmentation performance. Intermediate configurations between these two extremes are also possible. Figure~\ref{fig:perf-vs-eff} illustrates both modes for a multi-organ segmentation task.  

\vspace{0.2cm}
\noindent
\textbf{PGPS-Efficiency:}  
In PGPS-Efficiency, the batch size is kept constant throughout training. Since the patch size is smaller at early stages, the number of computations is reduced, leading to a substantial reduction in training time. This is illustrated in Figure~\ref{fig:perf-vs-eff} in green.

\vspace{0.2cm}
\noindent
\textbf{PGPS-Performance:}  
In PGPS-Performance, the GPU memory budget is fully utilized by dynamically increasing the batch size when smaller patches are used. This design leverages available resources more effectively, with the goal of achieving the highest possible segmentation performance. This is illustrated in Figure~\ref{fig:perf-vs-eff} in yellow.

\subsection{nnU-Net Framework Implementation}
\label{sec:curriculum-modes}

In the following section, we describe the integration details of default constant patch size sampling and the two different proposed PGPS sampling modes into the nnU-Net framework~\cite{isensee2021nnu}, which is the state-of-the-art 3D medical image segmentation framework~\cite{isensee2024nnu}.

\subsubsection{Baseline: Constant Patch Size Sampling}
The baseline sampling method utilized in this study is fixed or constant patch size (CPS) sampling. This sampling strategy corresponds to the conventional configuration of training a (patch-based) segmentation network. 

We have chosen nnU-Net~\cite{isensee2021nnu} as the underlying framework, as it provides state-of-the-art performance and is auto-configuring to downstream tasks. The default nnU-Net configuration is designed to achieve maximum global context by employing the largest possible patch size that fits within nnU-Net's typical GPU budget of 8GB. 

Given a patch-based setting, where the maximal processable patch size is smaller than the target volume size, nnU-Net will by default apply a batch size of two. In this setting, two patients are randomly sampled, while from the first patient, a foreground (FG) patch will be created, and from the second patient, a random patch will be sampled. This default 50\% FG patch ratio for a batch is designed to improve robustness in the presence of class imbalance~\cite{isensee2021nnu}. If multiple FG classes are present, one class will be picked randomly with equal probabilities. If nnU-Net sets a batch size larger than two, the framework employs a FG patch ratio of 33\%.

\subsubsection{PGPS Curricula}

We integrate the proposed Progressive Growing of Patch Size curricula within the nnU-Net framework. This integration involves maintaining the default nnU-Net settings while only adjusting the patch size for PGPS-Efficiency and PGPS-Performance, while also increasing the batch size for PGPS-Performance.

For the PGPS curricula, we have to define the minimal processable patch size and, given the maximal patch size, find all processable intermediate patch sizes that the network can process. In the case of nnU-Net, which applies a fully convolutional UNet backbone, the smallest processable patch size is dependent on the number of downsampling operations. The minimum patch size length depends on the number of downsampling operations for a UNet, and is given by $2^{num\_pool}$, where $num\_pool$ is the number of pooling operations for each axis. All intermediate patch sizes have to be fully divisible by $2^{num\_pool}$. The maximal patch size is set to the patch size nnU-Net would by default apply to the given task.

The patch size is increased in a step-wise linear fashion from the minimal possible patch size that the UNet allows until the patch size reaches the nnU-Net's target patch size. To have a smaller input tensor size growth, we only increase the patch size for one axis per step. We always select the axis with the lowest numerical value to increase the patch size. This scheme results in a higher average batch size during training, compared to naively sequentially increasing each axis. We train each patch size stage with the same number of epochs.

\vspace{0.2cm}
\noindent
\textbf{PGPS-Efficiency:}
For PGPS-Efficiency, only the patch size is adjusted during training, while the batch size is kept constant, using the default nnU-Net batch size. This results in reduced GPU memory utilization for smaller patch sizes.

\vspace{0.2cm}
\noindent
\textbf{PGPS-Performance:}
In PGPS-Performance, both patch size and batch size are adjusted to fully utilize the available GPU memory budget. By default, nnU-Net enforces a foreground patch ratio of 33\% for batch sizes larger than two, but switches to 50\% when the batch size is two. This setting would introduce a discontinuity in the class balance trajectory. To ensure a smooth progression, we instead fix the foreground patch ratio to 50\%, regardless of batch size. As there are different options in how to increase the batch size, we evaluated multiple batch construction strategies regarding efficiency and segmentation performance in~\ref{app:batch_strategies}. We identified that creating all patches from two patients per batch, using one patient for foreground patches and the other for random patches, performs best in terms of efficiency and segmentation performance. This allows for the creation of overlapping patches within the same patient. We follow this batching strategy throughout the manuscript.

\section{Experiments}

In Section~\ref{sec:datasets}, we introduce a set of widely used public medical image segmentation datasets used for the experiments. In Section~\ref{sec:benchmarking}, we evaluate the PGPS curriculum modes, analyze segmentation performance, computational costs (runtime and FLOPs), and convergence properties, and compare them to the constant patch size baseline. We present a correlation analysis between performance differences and task-specific characteristics in Section~\ref{sec:task-analysis}. A comparison between Progressive Resolution~\cite{tan2021efficientnetv2} and our proposed method is provided in Section~\ref{sec:prog-res}. In Section~\ref{sec:architectures}, the generalization of PGPS to different vision backbones is investigated. Finally, the stochastic training variability of the model performance across repeated trainings within the three sampling strategies is examined in Section~\ref{sec:stochastic-variability}.

\subsection{Dataset Descriptions}
\label{sec:datasets}
To test the curriculum methods, a wide range of publicly available segmentation datasets is utilized. In total, 15 different 3D medical semantic segmentation datasets have been gathered. The dataset mix includes the Medical Segmentation Decathlon~\cite{antonelli2022medical}, comprising 10 distinct datasets covering lesion, organ, and vessel tasks. Additionally, the ToothFairy2 dataset~\cite{Bolelli_2025_CVPR}, TotalSegmentatorV2~\cite{wasserthal2023totalsegmentator}, KiTS23~\cite{heller2024kidney}, AMOS22 CT/MRI task~\cite{ji2022amos} and BTCV~\cite{Landman2015MALBCV} are included. These datasets encompass a wide range of segmentation tasks, including lesions, various organs, bones, muscles, teeth, implants, and nerves. Furthermore, the datasets represent different 3D modalities such as CT, Conebeam-CT, and MRI, and vary in size from small to large, ranging from 20 to 1200 samples. They present a diverse range of class numbers, from 2 (binary) to 117 classes.

\subsection{Benchmarking of Methods on 15 Datasets}
\label{sec:benchmarking}
We evaluate segmentation performance, training runtime, and computational cost (FLOPs) of the PGPS curricula relative to default constant patch size (CPS) training. The default nnU-Net preprocessing and training pipeline is applied, using the 3D high-resolution patch-based variant for all tasks. For ToothFairy2 and TotalSegmentatorV2, mirroring data augmentation is disabled, as prior studies have demonstrated improved results without mirroring~\cite{isensee2024scaling, wasserthal2023totalsegmentator}. For both datasets, we also evaluate the segmentation performance for nnU-Net training without mirroring data augmentation in~\ref{app:orientation_sensitivity}.

\subsubsection{Segmentation Performance}  
To benchmark CPS, PGPS-Efficiency, and PGPS-Performance, we follow the standard nnU-Net evaluation procedure: models are trained using 5-fold cross-validation, and Dice scores are averaged across all foreground classes.  

To assess whether segmentation performance differs significantly between CPS and the PGPS curricula across the 15 datasets, paired Wilcoxon signed-rank tests are performed on the average foreground Dice scores. For this, we built one sample per dataset by pairing curriculum and CPS Dice score.  

\subsubsection{Training Runtime Tracking}  
Training time is monitored, excluding validation. Since experiments have been conducted on a cluster with heterogeneous GPU and CPU configurations, we propose a virtual relative runtime to enable fair comparisons. For the PGPS curricula, time per epoch was recorded. We then compute the average runtime for all maximal patch size epochs, which then represents the runtime of a single CPS epoch. Given that, we can then compute the virtual runtime of a full CPS training under the same hardware settings. We report the median over all 5-folds for the runtime.

\subsubsection{FLOPs Tracking}  
The number of Floating Point Operations (FLOPs) is directly proportional to the size of input tensors. We have recorded the total number of iterated voxels during training for each strategy and normalized them to the CPS value, yielding the relative difference in computational cost between PGPS curricula and CPS.

\subsubsection{Convergence Analysis}  
We analyze the convergence behavior of the three sampling strategies. For each strategy, models are trained across all 15 datasets with varying training lengths: 1\%, 10\%, 25\%, 50\%, and 100\% of total training iterations. This is done by reducing the number of iterations per epoch relative to the default 250 training iterations for nnU-Net. Evaluation at each training length is performed using 5-fold cross-validation.

\subsection{Task Characteristics Analysis} 
\label{sec:task-analysis}

To identify dataset characteristics associated with segmentation performance improvements, we correlate relative performance gains with the following task measures: (1) number of semantic classes, (2) training dataset size, (3) patch-to-volume coverage (the proportion of a full volume seen in a single patch), and (4) class imbalance, measured by the frequency of the smallest class.  

Spearman correlation tests are conducted for each task characteristic against the relative Dice score improvement between CPS and PGPS curricula. Separate tests are performed for each training length (1\%, 10\%, 25\%, 50\%, and 100\%).  

Additionally, we record the foreground ratio of input tensors, the number of unique classes seen per iteration, and the patch size ratio across all three sampling strategies.

\subsection{Progressive Resolution Curriculum}
\label{sec:prog-res}

Another input-length curriculum in computer vision is progressive resizing, also called Progressive Resolution~\cite{bolya2025perceptionencoderbestvisual, tan2021efficientnetv2}, which gradually increases input resolution from low to high. This approach also modifies input length, transitioning from small to large input tensors. To compare our PGPS curricula with Progressive Resolution, we evaluated both on BTCV, AMOS22, KiTS23, and MSD Lung Cancer, covering two highly imbalanced lesion tasks and two multi-organ tasks.  

We implement the Progressive Resolution curriculum within the nnU-Net framework as follows: Input tensor sizes are matched to those used for the PGPS curricula; however, instead of adjusting crop sizes, the default nnU-Net patch size is resampled to fit PGPS' target tensor size for each phase. Increasing the batch size, as done in PGPS-Performance, is computationally to expensive within nnU-Net for Progressive Resolution; therefore, we only evaluate Progressive Resolution using a fixed batch size. Segmentation performance of both curricula is compared using Dice scores in a 5-fold cross-validation.  

\begin{table*}[th!]
\centering
\caption{
Performance of CPS, PGPS-Efficiency, and PGPS-Performance on 15 diverse 3D Medical Image Segmentation tasks. CPS refers to constant patch size training, which is the standard nnU-Net training. All models have been trained from scratch. \textbf{Dice score [\%]:} evaluated in 5-fold cross-validation as in~\cite{isensee2019automated}; \textbf{$\blacktriangle x\%$ Rel. Dice score}: Relative Dice score differences to CPS (\gtr{} increase, \rtr{} decrease).
\textbf{Rel. Runtime [\%]:} relative runtime normalized to CPS's runtime; 
\textbf{Rel. FLOPs [\%]:} relative count of floating point operations to CPS value. \textbf{Bold}: Best performing sampling. \underline{Underlined}: Second best performing sampling.
}
\label{tab:MSD-performance}
\begin{small}
\begin{tabular}{l r r r r r r r r r}
\toprule
\multirow{2}{*}{\textbf{Dataset}} 
 & \multicolumn{3}{c}{\textbf{Dice score [\%]} $\uparrow$} 
 & \multicolumn{2}{c}{\textbf{$\blacktriangle x\%$ Rel. Dice score $\uparrow$}} 
 & \multicolumn{2}{c}{\textbf{Rel. Runtime [\%] $\downarrow$}} 
 & \multicolumn{2}{c}{\textbf{Rel. FLOPs [\%] $\downarrow$}} \\
\cmidrule(lr){2-4} \cmidrule(lr){5-6} \cmidrule(lr){7-8} \cmidrule(lr){9-10}
 & CPS & PGPS-Eff & PGPS-Perf 
 & PGPS-Eff & PGPS-Perf  
 & PGPS-Eff & PGPS-Perf  
 & PGPS-Eff & PGPS-Perf  \\
\midrule
MSD Brain         & 74.12 & \textbf{74.29} & \underline{74.15} & \gtr{0.23} & \gtr{0.04} & 42  & 83  & 34 & 84 \\
MSD Heart         & \underline{93.29} & 93.24 & \textbf{93.29} & \rtr{0.07} & \gtr{0.00} & 40  & 88  & 31 & 89 \\
MSD Liver         & 78.74 & \underline{78.76} & \textbf{80.60} & \gtr{0.03} & \gtr{2.36} & 51  & 85  & 38 & 84 \\
MSD Hippocampus   & 88.96 & \underline{89.12} & \textbf{89.15} & \gtr{0.20} & \gtr{0.22} & 67  & 112 & 33 & 92 \\
MSD Prostate      & 73.13 & \underline{75.26} & \textbf{76.31} & \gtr{2.91} & \gtr{4.35} & 43  & 87  & 30 & 87 \\
MSD Lung          & 70.00 & \underline{72.30} & \textbf{72.77} & \gtr{3.29} & \gtr{3.96} & 41  & 82  & 31 & 89 \\
MSD Pancreas      & \underline{68.68} & 68.60 & \textbf{68.80} & \rtr{0.12} & \gtr{0.17} & 39  & 87  & 31 & 88 \\
MSD Hepatic Vessel& \underline{68.61} & 68.05 & \textbf{68.98} & \rtr{0.82} & \gtr{0.54} & 47  & 87  & 33 & 88 \\
MSD Spleen        & \underline{97.02} & 95.85 & \textbf{97.15} & \rtr{1.21} & \gtr{0.13} & 42  & 86  & 33 & 89 \\
MSD Colon         & 48.41 & \underline{50.83} & \textbf{51.02} & \gtr{5.00} & \gtr{5.39} & 38  & 87  & 28 & 90 \\
BTCV              & \underline{83.37} & 83.05 & \textbf{83.81} & \rtr{0.38} & \gtr{0.53} & 41  & 88  & 29 & 89 \\
KiTS23            & 86.02 & \textbf{87.22} & \underline{86.46} & \gtr{1.40} & \gtr{0.51} & 45  & 86  & 27 & 84 \\
AMOS22            & \underline{88.62} & 88.10 & \textbf{88.78} & \rtr{0.59} & \gtr{0.18} & 49  & 87  & 33 & 89 \\
ToothFairy2       & 76.92 & \textbf{77.15} & \underline{77.00} & \gtr{0.31} & \gtr{0.11} & 34  & 85  & 26 & 88 \\
TotalSegmentatorV2        & \underline{87.82} & 85.09 & \textbf{88.15} & \rtr{3.11} & \gtr{0.38} & 33  & 108 & 30 & 90 \\
\midrule
\textbf{Norm. Avg.} 
 & 100.00 & \underline{100.47} & \textbf{101.26} 
 & \gtr{0.47} & \gtr{1.26} 
 & $44 \pm 9.5$ & $89 \pm 8.7$ 
 & $33 \pm 2.6$ & $88 \pm 2.4$ \\
\bottomrule
\end{tabular}
\end{small}
\end{table*}

\subsection{Different Architectures}
\label{sec:architectures}

The proposed curriculum can be applied to any vision backbone that supports flexible input sizes. To assess its applicability beyond CNNs, we further evaluated the curriculum on transformer-based hybrid architectures, specifically UNETR~\cite{hatamizadeh2022unetr} and SwinUNETR~\cite{hatamizadeh2021swin}, both combining a transformer encoder with a CNN decoder.

In transformer-based models, architectural constraints define the minimal feasible patch size and increment: for UNETR, this is determined by the internal patchify size (distinct from the patch size used in patch-based training), while for SwinUNETR it is defined by the Swin window size. As transformers typically rely on fixed input dimensions due to positional embeddings, we interpolate the positional embeddings for UNETR to match the varying patch sizes as in~\cite{dosovitskiy2020image}. SwinUNETR does by default not use any positional embeddings.

We integrate both architectures into the nnU-Net framework, train following the standard nnU-Net training, and evaluate on the BTCV dataset using a 5-fold cross-validation. All backbones are trained from scratch.

\subsection{Stochastic Training Variability}  
\label{sec:stochastic-variability}
Neural network training is inherently stochastic, a property that is amplified in patch-based pipelines, where both the sampled volume and patch location are randomly selected (within oversampling constraints). Consequently, repeated training runs can yield different models and segmentation performance.  

To quantify this variability, we repeat training on a single fold of the 5-fold cross-validation five times for each strategy. Experiments are conducted on the BTCV multi-organ segmentation dataset and the highly class-imbalanced MSD Lung Tumor dataset. Training lengths of 1\%, 10\%, and 100\% of the default iteration count were analyzed.  

To assess the stochastic impact of sampling, we build triplets of each strategy's segmentation performance outcome and evaluate all 125 possible outcome combinations per training length. As each single outcome could be a potential result, we are interested in which sampling strategy ranks best in each possible comparison. Thus, we count the number of scenarios in which each strategy outperformed both other strategies.

\section{Results}


\subsection{Benchmarking of Methods on 15 Datasets}

We evaluate the PGPS curricula relative to the conventional constant patch size (CPS) baseline across 15 popular 3D medical image segmentation tasks. Key measures include segmentation performance, training runtime, and computational cost in terms of FLOPs. Additionally, we also analyze the training convergence.

\subsubsection{Segmentation Performance}  
\label{subsubsec:perf}

Table~\ref{tab:MSD-performance} summarizes the performance benchmarking of the baseline compared to the curricula approaches on 15 datasets. Across the 15 datasets, PGPS-Performance achieves the highest average Dice score, yielding a relative improvement of 1.26\% over CPS. It outperforms CPS in every task, while ranking as the best-performing strategy in 12 of the 15 tasks. In all 15 datasets, PGPS-Performance is among the top two strategies, and only in the three tasks, KiTS23, ToothFairy2, and MSD Brain Tumor, PGPS-Performance is surpassed by PGPS-Efficiency.

PGPS-Efficiency also delivers modest overall performance gains, with a relative improvement of 0.47\% over all 15 tasks compared to CPS. It outperforms CPS in 8 tasks and is leading in 3 tasks. However, in 7 tasks, it has the lowest Dice score.

A two-sided paired Wilcoxon signed-rank test comparing PGPS-Performance to CPS across all 15 datasets reveals a significant improvement for PGPS-Performance ($p \approx  0.0001$). Additionally, PGPS-Performance significantly outperforms PGPS-Efficiency ($p \approx  0.0103$), confirming it as the superior strategy in terms of segmentation performance.  

\begin{figure*}[th!]
\centering
\includegraphics{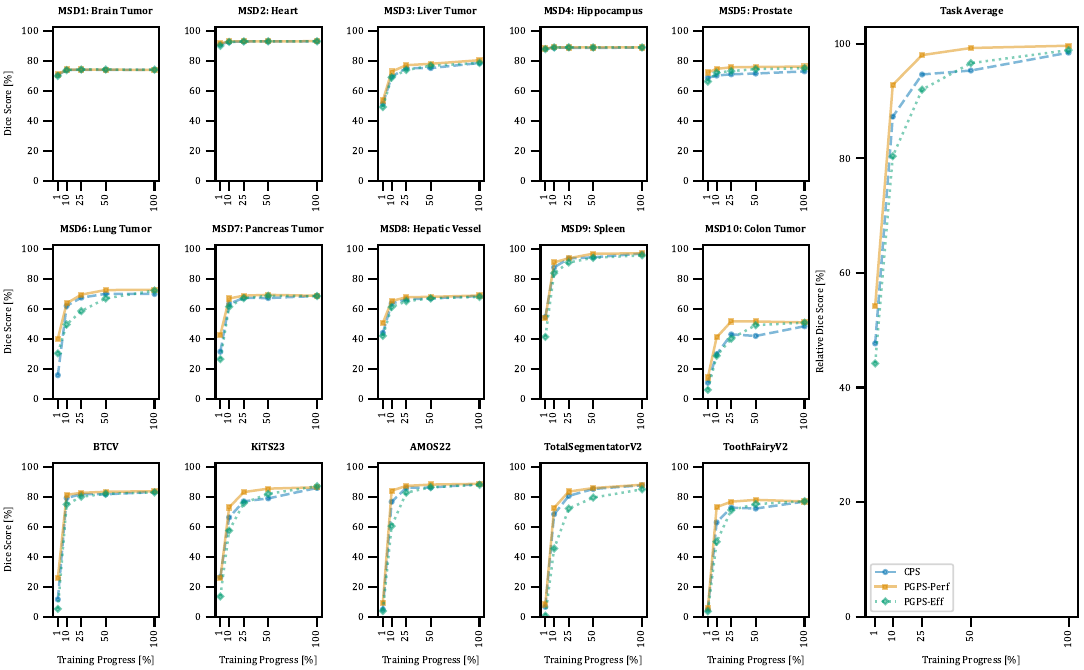}
\caption{Segmentation performance convergence of CPS, PGPS-Performance, and PGPS-Efficiency across training iterations. Dice scores are tracked across 15 segmentation tasks for different training lengths (1\%, 10\%, 25\%, 50\%, and 100\% of nnU-Net's default total training iterations). On average, PGPS-Performance exhibits the fastest convergence, while PGPS-Efficiency converges more slowly, due to the smaller input tensors that result in superior training speed. The final performance is, on average, best for PGPS-Performance, followed by PGPS-Efficiency and CPS. The average convergence over all 15 datasets (right) is computed by normalizing each task by its maximum performance and then averaging across all tasks.}
\label{convergence_over_iterations}
\end{figure*}

A two-sided paired Wilcoxon signed-rank test between PGPS-Efficiency and CPS shows no significant difference ($p \approx  0.6788$), indicating comparable segmentation performance between these two strategies.

\subsubsection{Training Runtime Tracking}  

Table~\ref{tab:MSD-performance} shows the relative runtime for all 15 datasets. PGPS-Performance requires approximately 89\% of CPS's runtime, ranging from 82\% to 112\% across tasks. Tasks for which PGPS-Performance yields longer training times than CPS, i.e. MSD Hippocampus and TotalSegmentatorV2, have the two largest start batch sizes of 1575 and 784, respectively. In total, the 15 tasks have an average batch size of around 399, ranging from 128 to 1575. PGPS-Efficiency substantially reduced training time, requiring only about 44\% of CPS runtime (range: 34\%–67\%), achieving the fastest training while maintaining comparable segmentation performance.

\subsubsection{FLOPs Tracking}  

As shown in Table~\ref{tab:MSD-performance}, PGPS-Efficiency utilizes only about 33\% of the FLOPs of CPS on average, with a range of 26\%–38\%. Although PGPS-Performance is designed to maximize the used GPU memory, it requires slightly fewer FLOPs than CPS, averaging 88\% (range: 84\%–92\%). The reduced computational cost arises from the discrete nature of batch and patch sizes, which prevents full usage of the GPU budget in every epoch, resulting in lower FLOPs.

\subsubsection{Convergence Analysis}

Figure~\ref{convergence_over_iterations} shows the convergence of segmentation performance across all 15 tasks over the number of iterations.

At 1\% training length, PGPS-Performance already achieves the highest Dice score in 13 tasks, ranking second to CPS only in MSD Spleen and KiTS23. At 10\% and 25\%, PGPS-Performance outperformed both other strategies across every task. At 50\%, PGPS-Performance outperforms both other strategies except for MSD Heart. However, the minimal differences in the results across methods and between training at 50\% and 100\% suggest that the performance is saturated in the MSD Heart task.

\begin{figure*}[th!]
\centering
\includegraphics{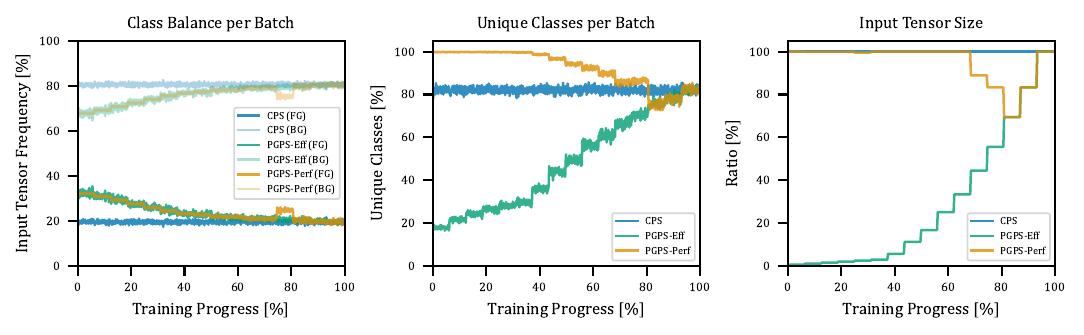}
\caption{Training characteristics of PGPS curricula for the BTCV dataset with 14 classes. \textbf{Left}: PGPS curricula improve foreground-background voxel balance compared to CPS due to on average smaller patch sizes. \textbf{Center}: PGPS-Performance achieves the highest average number of unique classes per batch/iteration due to larger batch sizes, while PGPS-Efficiency has the lowest. \textbf{Right}: CPS maintains a constant input tensor size; PGPS-Efficiency increases tensor size exponentially, while PGPS-Performance exhibits occasional size drops due to the discrete nature of patch and batch sizes.}
\label{training_voxel_metrics_combined_plots_onerow_with_bg_avg_unique_per_patch}
\end{figure*}

Overall, PGPS-Performance consistently converges faster than both CPS and PGPS-Efficiency, achieving superior results in the majority of training-length scenarios for each task.

In contrast, PGPS-Efficiency exhibits slower convergence than both other sampling strategies, as shown in Figure~\ref{convergence_over_iterations}, but requires substantially fewer computations, as seen in Table~\ref{tab:MSD-performance}, processing only about one-third of the voxels. Nevertheless, over the course of full training, PGPS-Efficiency reaches comparable, and in some cases even superior segmentation performance compared to CPS.

\subsection{Task Characteristics Analysis}
\label{task_characteristics_analysis}

To better understand the improved convergence for PGPS curricula, we track several task characteristics during model training for the BTCV dataset. Tracked characteristics include foreground class balance, number of unique classes per iteration, and patch size ratio. These are illustrated in Figure~\ref{training_voxel_metrics_combined_plots_onerow_with_bg_avg_unique_per_patch}. We find that both PGPS curricula exhibit higher class balance during early training phases compared to CPS, which maintains at a stationary foreground-to-background ratio. The reason is that fewer surrounding background voxels are contained in smaller foreground patches. At the final stage, PGPS curricula converge to the same class balance as CPS due to identical patch sizes. The standard deviation of class balance is larger for PGPS-Efficiency than for PGPS-Performance. PGPS-Performance shows a discontinuous drop in class balance during later training stages, attributed to the discrete nature of batch and patch sizes.  

Tracking the number of unique classes per iteration for the 14-class BTCV task, PGPS-Performance produces batches containing nearly all semantic classes ($\sim100\%$) during early patch size stages, converging to the stationary CPS value ($\sim82\%$) in later stages. Small drops in unique classes per iteration are again observed due to the discrete batch and patch sizes. PGPS-Efficiency begins with the smallest fraction of unique classes per iteration ($\sim18\%$), which gradually increases across patch size stages, eventually converging to the CPS value.  
The input tensor sizes of PGPS-Performance are the same as for CPS, but have occasional drops due to the discrete nature of batch and patch size. These drops result in the shorter training runtime observed for PGPS-Performance relative to CPS. PGPS-Efficiency exhibits a monotonic, stepwise exponential increase in patch size ratio.  

\vspace{0.2cm}
\noindent
\textbf{PGPS-Performance:}
Spearman tests do not yield any significant correlation between PGPS-Performance relative improvements and the dataset characteristics of the number of semantic classes, dataset size, or patch-to-volume coverage. The only significant correlation is negative, occurring for class imbalance measured via the smallest class frequency at training lengths of 1\% ($\rho \approx -0.7429$, $p \approx 0.0015$) and 10\% ($\rho \approx -0.5464$, $p \approx 0.0351$). This indicates that PGPS-Performance is particularly beneficial for highly imbalanced tasks.  

\vspace{0.2cm}
\noindent
\textbf{PGPS-Efficiency:} A significant positive correlation is observed between class imbalance (smallest class frequency) and relative performance at training lengths of 1\% ($\rho \approx 0.5321$, $p \approx 0.0412$), 10\% ($\rho \approx 0.7679$, $p \approx 0.0008$), and 25\% ($\rho \approx 0.6071$, $p \approx 0.0134$). Highly imbalanced tasks show reduced performance in short training scenarios, but PGPS-Efficiency eventually outperforms CPS at full training. This trend is evident for lesion tasks such as MSD Brain, MSD Liver Tumor, MSD Prostate, MSD Lung Tumor, MSD Colon, and KiTS23. Additionally, patch-to-volume coverage correlates significantly with relative performance at training lengths of 10\% ($\rho \approx 0.5157$, $p \approx 0.0491$) and 25\% ($\rho\approx 0.5318$, $p \approx 0.0382$).

\begin{table}[b!]
\centering
\caption{Dice scores of Progressive Resolution (Prog. Res.) and Progressive Growing of Patch Size (PGPS) curricula. PGPS increases patch size during training, whereas Progressive Resolution increases patch resolution. PGPS-Efficiency consistently outperforms Progressive Resolution, with the largest gains observed in highly imbalanced lesion tasks. Across all datasets, Progressive Resolution is the lowest-performing strategy. Results are reported as mean Dice score (\%) of a 5-fold Cross-Validation. \textbf{Bold:} best-performing strategy.}
\label{tab:prog-res}
\begin{tabular}{lcc}
\toprule
\textbf{Dataset}  & \textbf{Prog. Res.} & \textbf{PGPS-Eff} \\
\midrule
MSD Lung Tumor & 69.42 & \textbf{72.30} \\
KiTS23         & 84.63 & \textbf{87.22} \\
BTCV           & 82.93 & \textbf{83.05} \\
AMOS22          & 87.98 & \textbf{88.10} \\
\bottomrule
\end{tabular}
\end{table}

\subsection{Progressive Resolution Curriculum}

\begin{figure*}[t!]
\centering
\includegraphics{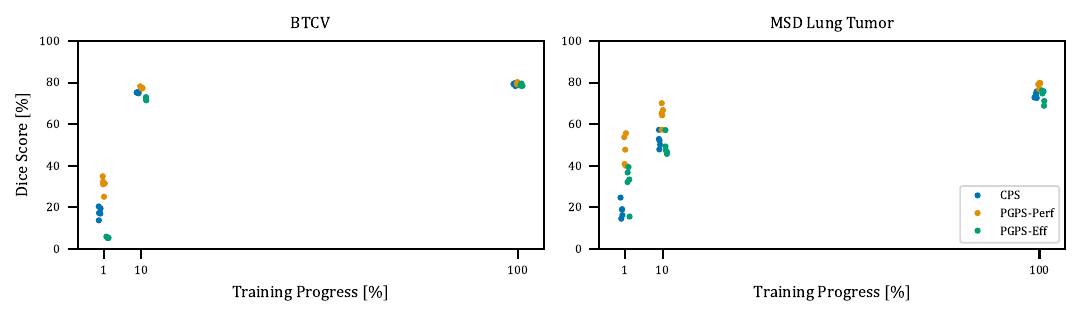}
\caption{Segmentation performance for repeated training of CPS, PGPS-Efficiency, and PGPS-Performance across different training lengths (1\%, 10\%, and 100\% of nnU-Net’s total training iterations). Each experiment is repeated five times on the same data split with different random seeds. Variability decreases with longer training, while mean performance increases. The highly class-imbalanced MSD Lung Tumor dataset exhibits higher variance than the multi-class BTCV dataset. PGPS-Performance shows improved convergence and less performance variability between repeated runs. PGPS-Efficiency results in slower convergence than CPS and PGPS-Performance and has the highest performance variability. PGPS-Performance outperforms both other strategies in 100\% of the 125 possible combinations for MSD Lung Tumor, while for BTCV, PGPS-Performance outperforms both other strategies in 58.4\% of all combinations, as well as CPS and 60\% of combinations.}
\label{fig:repeatability-combined}
\end{figure*}

We evaluate the Progressive Resolution curriculum, an input-length strategy widely applied in computer vision~\cite{tan2021efficientnetv2, wang2024efficienttrain++, bolya2025perceptionencoderbestvisual}, by adapting it to 3D patch-based medical image segmentation. Table~\ref{tab:prog-res} shows Dice scores for Progressive Resolution and PGPS-Efficiency. We find that our proposed PGPS-Efficiency consistently outperforms Progressive Resolution.

In the lesion segmentation task MSD Lung Tumor, Progressive Resolution lags behind PGPS-Efficiency by 2.9 Dice points (69.42\% vs.\ 72.30\%), and in KiTS23 by 1.2 points (84.63\% vs.\ 87.22\%). These results highlight that Progressive Resolution struggles in highly imbalanced settings, where PGPS-Efficiency provides a clear advantage.  

For multi-organ segmentation tasks, such as BTCV and AMOS22, the performance gap is smaller. Progressive Resolution achieves 82.93\% Dice on BTCV, close to PGPS-Efficiency (83.05\%). Similarly, in AMOS22, it reaches 87.98\%, compared to 88.10\% for PGPS-Efficiency. Although differences in these multi-organ tasks are less pronounced, Progressive Resolution is still the lowest-performing strategy for all cases.

\subsection{Different Architectures}
\begin{table}[b!]
\centering
\caption{Impact of PGPS-Efficiency (PGPS-Eff) and  PGPS-Performance (PGPS-Perf) on different backbones for the BTCV task. Results are reported as mean Dice score (\%) across five folds. PGPS-Performance consistently outperforms CPS across all architectures. PGPS-Efficiency converges for UNet and SwinUNETR but fails for UNETR due to gradient instability. Results are reported as mean Dice score (\%) of a 5-fold Cross-Validation. \textbf{Bold:} best-performing strategy. \underline{Underlined:} second-best-performing strategy.}
\label{tab:architectures}
\begin{tabular}{lccc}
\toprule
\textbf{Backbone} & \textbf{CPS} & \textbf{PGPS-Eff} & \textbf{PGPS-Perf} \\
\midrule
UNet (nnU-Net) \cite{isensee2021nnu} & \underline{83.37} & 83.05 & \textbf{83.81} \\
UNETR \cite{hatamizadeh2022unetr}    & \underline{71.20} & 0.00 & \textbf{72.76} \\
SwinUNETR \cite{hatamizadeh2021swin} & 71.62  & \underline{71.98} & \textbf{75.39}\\
\bottomrule
\end{tabular}
\end{table}

We further assess the generalizability of PGPS by applying it to transformer-based backbones (UNETR, SwinUNETR). Results for the BTCV dataset are reported in Table~\ref{tab:architectures} and compared to nnU-Net's default UNet backbone.

PGPS-Performance consistently improves segmentation performance across all architectures. Compared to CPS, Dice scores increase by $+0.44$ points for nnU-Net's fully convolutional UNet (83.37\% $\rightarrow$ 83.81\%), $+1.56$ points for UNETR (71.20\% $\rightarrow$ 72.76\%), and $+3.77$ points for SwinUNETR (71.62\% $\rightarrow$ 75.39\%). These results indicate that PGPS-Performance provides benefits not only for convolutional but also for transformer-based models.  

In contrast, PGPS-Efficiency yields mixed results. For nnU-Net, performance slightly decreases by $-0.32$ Dice points (83.37\% $\rightarrow$ 83.05\%), and for SwinUNETR it improves marginally by $+0.36$ points (71.62\% $\rightarrow$ 71.98\%). For UNETR, however, all training runs with PGPS-Efficiency diverge due to gradient explosion.

\subsection{Stochastic Training Variability}

Figure~\ref{fig:repeatability-combined} summarizes the variability in segmentation performance under repeated training. MSD Lung Tumor shows overall higher segmentation performance variability than for the BTCV dataset. Two consistent trends are observed: (i) outcome deviation per strategy decreases as training length increases, and (ii) sampling strategy PGPS-Performance dominates across both datasets and training length scenarios in segmentation performance over CPS and PGPS-Efficiency. 

\vspace{0.2cm}
\noindent
\textbf{BTCV:}  
PGPS-Performance dominates early training (1–10\%), achieving the highest segmentation performance in all comparisons against CPS and PGPS-Efficiency. At full training, it still achieves the majority of wins (58.4\%) over CPS and PGPS-Efficiency and the highest Dice score ($79.57\% \pm 0.37$), followed by CPS ($79.16\% \pm 0.47$) and PGPS-Efficiency ($78.85\% \pm 0.41$).

\vspace{0.2cm}
\noindent
\textbf{MSD Lung Tumor:} Across all training lengths, PGPS-Performance wins every sampling strategy comparison and achieves the highest Dice score ($78.98\% \pm 1.10$), outperforming CPS ($74.11\% \pm 1.23$) and PGPS-Efficiency ($73.34\% \pm 2.86$). Variability in segmentation performance per sampling strategy is larger at short training lengths but stabilizes as training progresses.

\section{Discussion}
\label{sec:discussion}

In this work, we introduced a novel curriculum learning strategy that progressively increases patch size during training. The curriculum was implemented in two modes: one optimized for segmentation performance and the other for computational efficiency. We have evaluated Dice score performance, training convergence, and computational cost across 15 diverse datasets covering lesion, multi-organ, vessel, muscle, and skeletal segmentation, and compared it to standard constant patch size (CPS) sampling. The performance mode demonstrates substantial Dice score improvements over CPS while simultaneously reducing computational costs in terms of FLOPs and training time. The efficiency mode matches the segmentation performance of CPS, while drastically cutting both FLOPs and training time by more than half. In addition, we have successfully applied the curriculum on multiple network backbones, including UNet, UNETR, and SwinUNETR, thereby confirming its broad applicability. In the following, we discuss the key factors driving these performance gains, as well as the limitations and implications of the Progressive Growing of Patch Size (PGPS) approach. We added a detailed comparison between the original proposed PGPS modes in \cite{fischer2024progressive} and the newly proposed modes in \ref{app:changes-pgps}. The newly proposed modes outperform the original modes in segmentation performance and computational efficiency. 

\subsection{Handling Class Imbalance}

Our proposed curriculum is a novel approach for tackling class imbalance in image segmentation. PGPS implicitly enhances class balance, as smaller patches contain proportionally fewer background voxels. This effect is particularly beneficial for imbalanced tasks, as class balance has shown a significant correlation with improved segmentation performance in our experiments. 

Another approach for improving class balance is Curriculum Adaptive Sampling for Extreme Data Imbalance (CASED)~\cite{jesson2017cased}, which explicitly increases foreground oversampling at the beginning of training and gradually reduces it over time. While CASED relies on a large batch size, it trades off patch size (e.g., patch size of $68^3$ and batch size of 16). Our method, however, follows the configuration principles of modern segmentation frameworks such as nnU-Net, which prioritize a large patch size over the batch size (typically batch size of 2). In principle, CASED could be integrated into PGPS to further enhance class balance, offering a promising direction for future research.

\subsection{Factors Contributing to Segmentation Performance Gain}
The segmentation performance gains achieved by PGPS curricula can be attributed to several factors beyond the improved class balance. For PGPS-Performance, the model is exposed to a larger number of unique classes per iteration, although the correlation between class count and performance was not statistically significant. Additionally, the larger batch sizes in PGPS-Performance increase exposure to data augmentations and promote sampling from more diverse patch locations, leading to broader coverage of the data distribution without losing spatial resolution or altering the forced foreground-to-background ratio. This is supported by the near-significant correlation between patch-to-volume coverage and performance improvement ($\rho \approx -0.4978$, $p \approx 0.0590$) at 50\% training length for PGPS-Performance. Interestingly, most correlations are strongest at shorter training lengths. We attribute this to performance saturation at full training, which dampens differences between sampling strategies. Performance gains are consistent with existing studies on LLM training, indicating that shorter input sequences reduce gradient variance, improving training stability~\cite{li2022stability}. 

Furthermore, we hypothesize that the aforementioned factors have contributed to the overall performance gain of PGPS over CPS training, but disentangling those factors will be subject to future research. While PGPS-Efficiency cut training runtime strongly, while keeping comparable segmentation performance to CPS, PGPS-Performance was able to significantly outperform standard CPS training on average on 15 datasets, improving segmentation performance on each dataset, while also slightly reducing computational costs.

\subsection{PGPS on Different Backbones}
PGPS-Performance improves the segmentation performance for all different tested backbones. Therefore, we hypothesize that limiting the contextual information available to the network, by training on smaller image patches, facilitates more stable and efficient learning, similar to findings reported for large language models in~\cite{li2022stability}. When trained from scratch, Transformer-based architectures generally lag behind CNNs due to their lack of strong inductive biases~\cite{touvron2021training}. While large-scale pretraining is a common solution to mitigate this limitation~\cite{dosovitskiy2020image}, several approaches have been proposed to introduce inductive bias directly into Transformers, such as CNN-based teacher distillation~\cite{touvron2021training} or architectural modifications like shifted window attention~\cite{lee2021vision,liu2021swin}.

Our results demonstrate that the proposed PGPS-Performance curriculum consistently improves segmentation performance across all evaluated architectures, including UNet, UNETR, and SwinUNETR. In contrast, the PGPS-Efficiency variant converges successfully for UNet and SwinUNETR but fails for UNETR. We hypothesize that the stronger inductive biases in UNet and SwinUNETR enable effective learning even under reduced input context, whereas UNETR appears more sensitive to limited information, leading to training instabilities. PGPS-Performance appears to enhance the training signal compared to constant patch size training, thereby improving Dice scores for UNETR as well. Another potential factor for UNETR’s instability under PGPS-Efficiency is that simple interpolation of positional embeddings may be suboptimal, while the other two architectures do not rely on such embeddings.

In summary, we hypothesize that training with PGPS-Performance facilitates the learning process by improving the quality of the training signal, particularly for architectures lacking strong inductive bias, such as Transformers. Although we did not explore pretrained Transformer weights in this work, an open research question remains how transfer learning and pretrained representations interact with PGPS-based curricula.

\subsection{Comparison to Progressive Resolution}
Our results demonstrate that PGPS consistently outperforms Progressive Resolution in the tested patch-based segmentation tasks. This is likely due to improved class balancing with PGPS, while Progressive Resolution does not affect the class balance and thereby matches the class balance of CPS. The substantial drop in performance of Progressive Resolution in highly imbalanced tasks further supports this claim. Progressive Resolution is thus only suited for global image-level prediction tasks such as classification~\cite{tan2021efficientnetv2}, image generation~\cite{karras2017progressive}, contrastive learning~\cite{koccyiugit2023accelerating}, and CLIP alignment~\cite{lireclip}. By contrast, PGPS is more effective for dense prediction tasks, such as segmentation and potentially object detection, particularly in class-imbalanced scenarios.

\subsection{Stochastic Training Variability and Model Validation}
\label{disc:stat-variance}
A further challenge is the variance in reported performances in repeated training of the same model due to stochastic training variability in deep learning. For example, Dice scores of the default nnU-Net on the BTCV task vary widely, even under the same data splits: 83.37\% (ours), 83.56\% (MedNeXt~\cite{roy2023mednext}), and 83.08\% (nnU-Net Revisited~\cite{isensee2024nnu}). Surprisingly, even the stronger and larger nnU-Net ResEnc variants (M: 83.31\%, L: 83.35\%, XL: 83.28\%) reported in~\cite{isensee2024nnu} are outperformed by smaller default nnU-Net models in MedNeXt~\cite{roy2023mednext} and in our experiments in Table~\ref{tab:MSD-performance}. While the extent to which sampling alone explains these discrepancies remains unclear, differences in preprocessing (e.g., different sample origin cache in nnU-Net) may also contribute. 

Furthermore, to assess the stochastic training variability, we report nnU-Net validation performance from related work that were trained on one of our benchmarking datasets, and compare them to our proposed PGPS framework in ~\ref{app:reported_nnunet_scores}. We have found that our proposed curriculum resulted in better-performing models than the scores reported in the literature, which were trained via conventional CPS sampling.

In~\cite{isensee2024nnu}, Isensee \textit{et al.} highlight that not all datasets are equally suitable for model comparison due to a high statistical variance (intra-method) or a low systematic (inter-method) variance. Suitable datasets, such as AMOS22, KiTS23, and LiTS, showed higher inter-method than intra-method standard deviation, whereas worse-suited datasets like BTCV showed lower standard deviation in their experiments. Improved convergence via PGPS-Performance yields reduced stochastic difference due to sampling, leading to improved segmentation performance and lower segmentation performance variance as seen in experiment~\ref{sec:stochastic-variability}. This provides a better signal-to-noise ratio, making PGPS-Performance a more reliable strategy for fair model comparison than standard CPS. 

Following the strategy in~\cite{isensee2024nnu}, we exemplarily compute inter- and intra-method standard deviation for PGPS-Performance for BTCV and KiTS23. Details on the experiment and results are reported in~\ref{app:resenc_variants}. We found that PGPS-Performance sampling yields better method comparison, as the inter-method to intra-method standard deviation ratio is improved for both datasets when switching from CPS to PGPS-Performance sampling.

\section{Conclusion and Outlook}
\label{sec:conclusion}

In this work, we have introduced a novel curriculum for semantic segmentation based on increasing the patch size during training. We have evaluated the curriculum on 15 diverse and popular 3D medical image segmentation datasets against the default constant patch size baseline. We have proposed two curriculum modes: \textbf{PGPS-Performance}, which has been shown to consistently outperform constant patch size sampling on all 15 datasets, in terms of Dice score, while requiring only 90\% of the original training time, and \textbf{PGPS-Efficiency}, which matches the performance of conventional constant patch size training while drastically reducing training time to 44\%. We have demonstrated that modifying the sampling strategy substantially accelerates training convergence and statistically significantly improves final Dice score performance for patch-based 3D medical image segmentation. 

We have shown that Progressive Resolution, a well-known input-length automatic curriculum based on image resolution~\cite{tan2021efficientnetv2, bolya2025perceptionencoderbestvisual, lireclip, koccyiugit2023accelerating}, leads to performance decreases in our experiments and is more suited for solving global image-level tasks like classification or contrastive pretraining. This underlines the need and research gap for a curriculum learning approach for dense prediction tasks. In contrast, our proposed curriculum based on patch size did improve the segmentation performance and is, to the best of our knowledge, the first input-length curriculum based on patch size in the computer vision and medical imaging domains.

Our analysis shows that tasks with strong class imbalance benefit most from PGPS, underscoring the importance of sampling strategies in patch-based training. Importantly, the advantages of PGPS are not limited to UNet-style architectures. We have been able to demonstrate consistent improvements for PGPS-Performance across both convolutional and transformer-based models, including UNet, UNETR, and SwinUNETR, highlighting the broad applicability of the method.

A key strength of our approach is its simple integration into any segmentation backbone training. Depending on the application, users can choose between PGPS-Efficiency for rapid experimentation with reduced compute and carbon footprint, or PGPS-Performance for maximizing segmentation performance in deployment settings, while also benefiting from a reduced carbon footprint. Furthermore, PGPS-Performance enables more reliable model comparison by improving the signal-to-noise ratio, due to improved convergence, and thereby providing a clearer signal for methodological benchmarking.\\

We envisage several promising directions for further research. The curriculum approach sequence length warm-up was shown to reduce the gradient variance in training large language models, enabling larger learning rates and batch sizes, resulting in extremely short training time to only 6\% of the original time~\cite{li2022stability}. We have not explored hyperparameter tuning in our experiments, but anticipate that this could lead to further runtime reduction.

While most research in medical image segmentation is predominantly conducted in the architectural design of backbones, we argue that the sampling strategy itself is a vastly overlooked area and requires more exploration. Advances in differentiable and adaptive sampling strategies, such as differentiable top-$k$ Patch Sampling~\cite{jeon2025no}, open new directions for optimizing patch-based sampling. Beyond patch sampling, alternative paradigms such as multi-scale frameworks~\cite{KAMNITSAS201761} or cascaded architectures~\cite{isensee2021nnu} aim to reduce reliance on patching altogether, but patch-based methods remain dominant in segmentation performance today~\cite{jeon2025no}.

\textbf{In summary, we recommend PGPS-Performance as the new default sampling strategy for training 3D patch-based segmentation backbones for deployment.} It is conceptually simple, easy to implement within existing frameworks such as nnU-Net, and provides consistent improvements in both segmentation performance and resource efficiency, establishing a strong novel sampling strategy in patch-based 3D medical image segmentation.

\section*{Acknowledgements}
SMF, JK and JAS acknowledge funding from Munich Center for Machine Learning (MCML). SMF, JCP, and JAS acknowledge funding by the Deutsche Forschungsgemeinschaft (DFG, German Research Foundation) – 515279324 / SPP 2177. SMF acknowledges funding from the EVUK program (”Next-generation AI for Integrated Diagnostics”) of the Free State of Bavaria. JK and JCP acknowledge funding from the Wilhelm Sander Foundation in Cancer Research (2022.032.1). JK acknowledges funding from the DAAD program Konrad Zuse Schools of Excellence in Reliable Artificial Intelligence, sponsored by the German Federal Ministry of Education and Research. RO and KL acknowledge funding from the EU Horizon Europe and Horizon 2020 research and innovation programme under grant agreement No 101057699 (RadioVal) and No 952103 (EuCanImage), respectively. RO acknowledges a research stay grant from the Helmholtz Information and Data Science Academy (HIDA). JAS and LD are supported by the German Federal Ministry of Research, Technology and Space (DECIPHER-M, 01KD2420G). JAS and DML acknowledge funding from HELMHOLTZ IMAGING, a platform of the Helmholtz Information \& Data Science Incubator.

\section*{Data Availability}
All data are publicly available. We provide detailed descriptions and proper citations for each dataset used to ensure full transparency and reproducibility. Additionally, all code for data preprocessing and preparation is publicly released and can be used to fully reproduce our results. Our code is publicly available at \url{https://github.com/compai-lab/2025-MedIA-fischer/releases/tag/v2.6.2}.

\section*{Conflicts of Interest}
The authors declare that they have no known conflict of interest nor competing financial interests or personal relationships that could have appeared to influence the work reported in this paper.

\section*{Declaration of generative AI and AI-assisted technologies in the manuscript preparation process}
During the preparation of this work the author(s) used ChatGPT in order to improve the readability of the manuscript. After using this tool, the author(s) reviewed and edited the content as needed and take(s) full responsibility for the content of the published article.

\section*{CRediT authorship contribution statement}
\textbf{Stefan M. Fischer:} Conceptualization, Methodology, Software, Writing - Original Draft
\textbf{Johannes Kiechle:} Validation, Writing - Review \& Editing
\textbf{Laura Daza:} Validation, Writing - Review \& Editing
\textbf{Lina Felsner:} Validation, Writing - Review \& Editing,
\textbf{Richard Osuala:} Validation, Writing - Review \& Editing,
\textbf{Daniel Lang:} Validation, Writing - Review \& Editing,
\textbf{Karim Lekadir:} Validation, Writing - Review \& Editing, Funding acquisition, 
\textbf{Jan C. Peeken:} Validation, Writing - Review \& Editing, Supervision, Funding acquisition,
\textbf{Julia A. Schnabel:} Validation, Writing - Review \& Editing, Supervision, Funding acquisition

\bibliographystyle{elsarticle-num}
\bibliography{medical_image_analysis_cleaned}

\clearpage
\appendix
\clearpage

\section{Batch Construction Strategies}
\label{app:batch_strategies}

\begin{table*}[th]
\centering
\caption{
Comparison of Dice scores, FLOPs, and relative runtime across three datasets 
for Constant Patch Size (CPS) and Progressive Growing of Patch Size (PGPS) 
variants with split- and single-crop sampling. 
All values are normalized to the CPS baseline. 
PGPS-Perf (Single-Crop) also improves Dice score over CPS but incurs substantial runtime overhead due to increased dataloading costs.
}
\label{tab:pgps_efficiency}
\renewcommand{\arraystretch}{1.1}
\setlength{\tabcolsep}{5pt}
\begin{small}
\begin{tabular}{lccccccccc}
\toprule
\textbf{Metric} & 
\multicolumn{3}{c}{\textbf{MSD6 Lung Tumor}} & 
\multicolumn{3}{c}{\textbf{KiTS23}} & 
\multicolumn{3}{c}{\textbf{AMOS22}} \\
\cmidrule(lr){2-4} \cmidrule(lr){5-7} \cmidrule(lr){8-10}
 & \textbf{CPS} & 
   \makecell[c]{\textbf{PGPS-Perf}\\(Split-Crop)} & 
   \makecell[c]{\textbf{PGPS-Perf}\\(Single-Crop)} & 
   \textbf{CPS} & 
   \makecell[c]{\textbf{PGPS-Perf}\\(Split-Crop)} & 
   \makecell[c]{\textbf{PGPS-Perf}\\(Single-Crop)} & 
   \textbf{CPS} & 
   \makecell[c]{\textbf{PGPS-Perf}\\(Split-Crop)} & 
   \makecell[c]{\textbf{PGPS-Perf}\\(Single-Crop)} \\
\midrule
\textbf{Dice score [\%]} $\uparrow$     & 70.00 & \textbf{72.77} & \underline{72.45} & 86.02 & \underline{86.46} & \textbf{87.18} & 88.62 & \textbf{88.78} & \underline{88.77} \\
\textbf{Rel. FLOPs [\%]} $\downarrow$   & 100   & \textbf{\underline{89}}    & \textbf{\underline{89}}    & 100   & \textbf{\underline{84}}    & \textbf{\underline{84}}    & 100   & \textbf{\underline{89}}    & \textbf{\underline{89}}    \\
\textbf{Rel. Runtime [\%]} $\downarrow$ & 100   & \textbf{82}    & \underline{87} & \underline{100}   & \textbf{86}    & 205 & \underline{100}   & \textbf{87} & 198 \\
\bottomrule
\end{tabular}
\end{small}
\end{table*}

\textbf{Setup:}  
To efficiently utilize GPU resources during early PGPS-Performance stages with smaller patches, the batch size was increased either by (1) drawing patches from more patient volumes or (2) extracting multiple patches per patient volume.  We investigated four patch sampling strategies, \textit{Single-Volume}, \textit{Multi-Crop}, \textit{Split-Crop}, and \textit{Single-Crop}, on the MSD Lung Tumor dataset to analyze their influence on convergence and generalization.

\textbf{Batching Strategies:}  
\textit{Single-Volume} crops all patches from one patient volume, while \textit{Multi-Crop} and \textit{Split-Crop} sample from two patient volumes. In \textit{Split-Crop}, one patient provides foreground patches and the other background patches, whereas \textit{Multi-Crop} draws both from each patient.  
\textit{Single-Crop} follows the nnU-Net scheme, loading as many patients as the batch size, drawing one patch per patient, thereby maximizing patient diversity, preventing patch overlaps.

\begin{figure}[b!]
\centering
\includegraphics{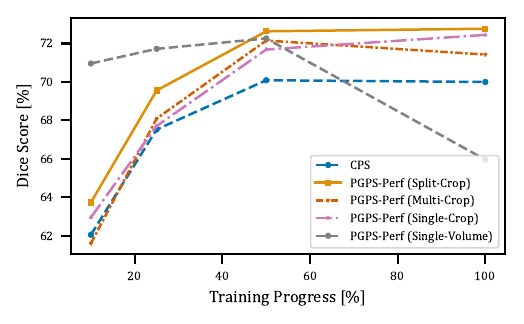}
\caption{Convergence of different batching strategies for PGPS-Performance on the MSD Lung Tumor dataset. Increasing batch size can be achieved by sampling from more patients or generating multiple crops per patient. Overfitting occurs when both foreground and background patches are drawn from the same patient.}
\label{fig:overfitting_Lung}
\end{figure}

\textbf{Results on MSD Lung:}  
All PGPS-Performance batching strategies converged faster than the baseline Constant Patch Size (CPS) setup (Fig.~\ref{fig:overfitting_Lung}). \textit{Single-Volume} exhibited strong overfitting, and \textit{Multi-Crop} showed moderate overfitting. \textit{Split-Crop} achieved the most stable convergence and highest Dice performance, while \textit{Single-Crop} slightly lagged in convergence speed.  
Runtime analysis showed that \textit{Split-Crop}\textit{Split-Crop} and \textit{Multi-Crop} ran efficiently (82\% of CPS runtime), whereas \textit{Single-Crop} required more dataloading time (87\% of CPS runtime) because for each batch, many distinct patient volumes have to be loaded.

\textbf{Cross-Dataset Runtime Comparison:}  
While \textit{Split-Crop} allows patch overlaps, \textit{Single-Crop} prevents this. To quantify the effect on large high-quality benchmark datasets, we pick  AMOS22 and KiTS23 to evaluate segmentation performance and computational costs. 
Table~\ref{tab:pgps_efficiency} reports Dice scores, FLOPs, and runtime normalized to the CPS baseline. Both PGPS variants consistently improved segmentation performance over CPS, but \textit{Split-Crop} was much more efficient. \textit{Single-Crop} runtime increased drastically, 205\% on KiTS23 and 198\% on AMOS22, caused by extensive I/O from loading large numbers of patient volumes. In contrast, \textit{Split-Crop} achieved nearly identical performance with far lower computational overhead (82–87\% of CPS runtime), while achieving higher segmentation performance than \textit{Single-Crop} in AMOS22 and MSD Lung. For larger-batch datasets such as TotalSegmentatorV2 (batch size 1575), these runtime penalties for \textit{Single-Crop} would become prohibitively large.  

\textbf{Conclusion:}  
The choice of batch construction has a strong impact on generalization and computational efficiency. \textit{Split-Crop} offers the best trade-off between convergence speed, segmentation performance, and runtime, and was therefore adopted as the default batching strategy for all PGPS-Performance experiments. The \textit{Single-Crop} strategy remains useful for applications prioritizing maximal patch variety, though it would benefit from optimized dataloading implementations. Enlarging patch diversity by preventing patch overlaps did not seem to improve performance in our experiments. 

Furthermore, we hypothesize that the low batch diversity is the main reason for overfitting. With batch diversity, we refer to the number of possible combinations between foreground and background patches of different patients. Given a dataset with $n$ samples, for \textit{Single-Volume} only $n$ combinations are possible, as we pair foreground and background of \textbf{Patient A}. This low batch diversity ends up in strong overfitting.

For \textit{Multi-Crop}, we pair the foreground and background of \textbf{Patient A} with those of \textbf{Patient B}. We thus build unordered pairs. For this strategy we only observe slight overfitting. 

For \textit{Split-Crop}, we pair the foreground of \textbf{Patient A} with the background of \textbf{Patient B}, which are ordered pairs and thus more as for \textit{Multi-Crop}. This batch diversity is the same as for standard nnU-Net CPS training (batch size of two). Here, we do not observe any overfitting.

Maximal variety is possible with \textit{Single-Crop}. As for each patch, a new patient is used, and thus, the maximal possible batch diversity. This large batch diversity did not improve performance over \textit{Split-Crop} as there was slower convergence. Longer training might yield better results for \textit{Single-Crop} than for \textit{Split-Crop}.

\section{Orientation Sensitivity}
\label{app:orientation_sensitivity}

\textbf{Setup:}  
Smaller patches reduce global spatial context, which can hinder segmentation of directionally defined structures (e.g., left/right or upper/lower). Orientation sensitivity varies across datasets: BTCV and AMOS22 each include 4 directionally defined classes out of 14 and 15, TotalSegmentatorV2 includes 66 out of 107, and ToothFairy2 includes 32 out of 42. Mirror augmentation, the nnU-Net default, can further obscure orientation cues by inverting anatomical structures.  
We trained CPS and PGPS-Performance with and without mirror augmentation on these datasets using 5-fold cross-validation.

\textbf{Results:}  
As shown in Table~\ref{tab:mirror_impact}, the influence of mirror augmentation depends strongly on dataset orientation sensitivity.  
For BTCV and AMOS22, effects were minimal, with PGPS-Performance consistently outperforming CPS.  
In TotalSegmentatorV2, disabling mirroring substantially improved results, and PGPS-Performance achieved the best overall Dice score.  

The strongest degradation occurred in ToothFairy2, where most classes are directionally defined; mirror augmentation severely reduced performance for CPS and PGPS-Performance, but particularly for the curriculum. Disabling mirroring did result in a large performance gain for both sampling strategies in ToothFairy2, where PGPS-Performance did outperform CPS.

Overall, PGPS-Performance remains beneficial when orientation-sensitive classes form only a subset of labels, but can suffer when such classes dominate the dataset. We therefore recommend caution with mirror augmentation in tasks with large numbers of directional-defined classes, as it may conflict with PGPS’s progressive curriculum and limit performance gains. Additionally, a CPS-trained network would also benefit from disabling mirroring.

\begin{table}[h!]
\centering
\caption{Impact of mirror augmentation on CPS and PGPS across datasets containing directionally defined semantic classes. Results are mean Dice scores (\%) over 5-fold cross-validation.}
\label{tab:mirror_impact}
\begin{tabular}{lcc|cc}
\toprule
\multirow{2}{*}{\textbf{Dataset}} 
& \multicolumn{2}{c|}{\textbf{With Mirror}} 
& \multicolumn{2}{c}{\textbf{Without Mirror}} \\
\cmidrule(lr){2-3} \cmidrule(lr){4-5}
& CPS & PGPS-Perf & CPS & PGPS-Perf \\
\midrule
BTCV              & 83.37 & \textbf{83.81} & 82.93 & \textbf{83.09} \\
AMOS22            & 88.62 & \textbf{88.78} & 88.56 & \textbf{88.63} \\
TotalSegV2        & 84.56 & \textbf{84.64} & 87.82 & \textbf{88.15} \\
ToothFairy2       & \textbf{63.23} & 47.08 & 76.92 & \textbf{77.00} \\
\bottomrule
\end{tabular}
\end{table}

\section{nnU-Net Performance in Related Work}
\label{app:reported_nnunet_scores}

\begin{table*}[th!]
\centering
\caption{
Comparison of nnU-Net baselines and related models across the Medical Segmentation Decathlon (MSD), BTCV, KiTS23, and AMOS22-CT/MRI datasets. 
Reported values denote mean Dice scores over 5-fold cross-validation as provided in the respective publications. 
All models, except the last row, were trained using the default \textbf{constant patch size (CPS)} strategy. 
The final row reports performance with our proposed \textbf{Progressive Growing of Patch Size (PGPS-Performance)} curriculum. 
Bold values indicate the best result per dataset. This overview is likely not exhaustive. Overall, Models trained via our proposed PGPS-Performance curriculum yield the best segmentation performance, delivering the highest performance on all datasets, except MSD2.
}
\label{tab:literature_comparison}
\setlength{\tabcolsep}{4pt}
\renewcommand{\arraystretch}{1.05}
\begin{scriptsize}
\begin{tabular}{lcccccccccccccc}
\toprule
\textbf{Original Publication} & MSD1 & MSD2 & MSD3 & MSD4 & MSD5 & MSD6 & MSD7 & MSD8 & MSD9 & MSD10 & BTCV & KiTS23 & AMOS22-CT/MRI \\
\midrule
\textbf{Constant Patch Size:} &&&&&&&&& \\
nnU-Net (orig.) \cite{isensee2019automated} & 74.11 & 93.28 & 79.71 & 88.91 & 75.37 & 72.11 & 67.45 & 68.37 & 96.38 & 45.53 & 82.79 & -- & -- \\
nnU-Net v2 (Repo) \cite{nnunet-github} & 73.98 & 93.34 & 79.53 & 88.95 & 75.01 & 69.83 & 67.17 & 68.41 & 96.90 & 45.04 & 83.25 & -- & -- \\
nnU-Net revisited \cite{isensee2024nnu}& -- & -- & -- & -- & -- & -- & -- & -- & -- & -- & 83.08 & 86.04 & 88.64 \\
MedNeXt \cite{roy2023mednext}& -- & -- & -- & -- & -- & -- & -- & -- & -- & -- & 83.56 & -- & -- \\
RecycleNet \cite{koehler2024recyclenet} & -- & -- & -- & -- & -- & -- & -- & -- & -- & -- & 82.96 & -- & 88.58 \\
Extending nnU-Net is All You Need \cite{isensee2023extending} & -- & -- & -- & -- & -- & -- & -- & -- & -- & -- & -- & -- & 88.64 \\
Auto-nnU-Net \cite{becktepe2025auto} & 73.98 &  \textbf{93.39} & 79.45 & 89.04 & 73.53 & 68.33 & 66.07 & 68.31 & 96.66 & 46.04 & -- & -- & -- \\
Ours (CPS) & 74.12 & 93.29 & 78.74 & 88.96 & 73.13 & 70.00 & 68.68 & 68.61 & 97.02 & 48.41 & 83.37 & 86.02 & 88.62 \\
\midrule
\textbf{Progressive Growing of Patch Size:} &&&&&&&&& \\
Ours (PGPS-Perf) & \textbf{74.15} & 93.29 & \textbf{80.60} & \textbf{89.15} & \textbf{76.31} & \textbf{72.77} & \textbf{68.80} & \textbf{68.98} & \textbf{97.15} & \textbf{51.02} & \textbf{83.81} & \textbf{86.46} & \textbf{88.78} \\
\bottomrule
\end{tabular}
\end{scriptsize}
\end{table*}

\textbf{Setup:} We assessed the variability in segmentation performance that was reported for nnU-Net baselines in the literature. Therefore, we reviewed publications that employed nnU-Net in their studies and reported results from 5-fold cross-validation on datasets contained in our benchmark.   

\textbf{Results:} In total, we identified nine sources (eight publications and the official DKFZ nnU-Netv2 GitHub repository) that provided relevant validation scores. Although this collection is not exhaustive, it reflects the diversity of nnU-Net results currently cited in the field. 

\begin{figure}[b!]
\centering
\includegraphics{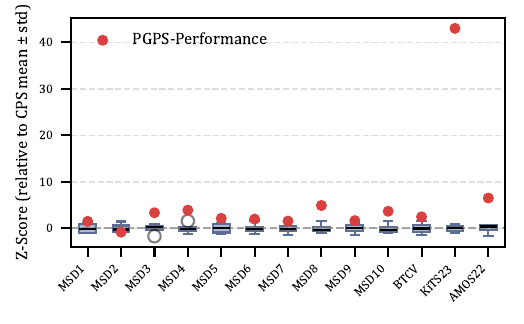}
\caption{Reported nnU-Net Segmentation Performances in Literature compared to our PGPS-Performance Results. We collected reported nnU-Net performances, trained via constant patch size (CPS), and plotted their normalized standard deviation in Dice score against instances trained via PGPS-Performance. Training with PGPS-Performance resulted in the best-performing models in all datasets, except for MSD2. For all datasets, we had 4 or more CPS performance values, while for KiTS23, we only had 2 values. Performance values per publication can be seen in~\ref{tab:literature_comparison}.}
\label{fig:reported_nnunet_performances}
\end{figure}

Table~\ref{tab:literature_comparison} summarizes Dice scores reported across the Medical Segmentation Decathlon (MSD), BTCV, KiTS23, and AMOS22 datasets. Most of these results originate from the DKFZ research group~\cite{roy2023mednext,isensee2019automated,isensee2024nnu,nnunet-github,koehler2024recyclenet, isensee2023extending}, who developed and maintain nnU-Net. Another external publication was found using nnU-Net as baseline~\cite{becktepe2025auto}. In Figure~\ref{fig:reported_nnunet_performances}, we plot the normalized standard deviation in performance over all found runs per dataset. Note that scores originate from different nnU-Net versions (nnU-Netv1 and nnU-Netv2), but DKFZ claims that both maintain equivalent segmentation performance~\cite{nnunet-github}.

For datasets, where we found multiple CPS-based results, as seen in Table~\ref{tab:literature_comparison}, the variance in reported performance underscores the variability of training outcomes across different runs. Despite this variability, our proposed PGPS-Performance curriculum consistently achieves the highest Dice score across all evaluated datasets except MSD2. 

The marginal exception in MSD2 could result from convergence saturation, where other factors, such as network initialization or cached class sampling origins, might play a stronger role than improved sampling.

Overall, these findings confirm that our proposed PGPS-Performance does result in improved performance over repeated runs on different hardware and environments.

\section{More Reliable Backbone Comparison}
\label{app:resenc_variants}

\textbf{Setup:} As discussed in Section~\ref{disc:stat-variance}, the proposed PGPS-Performance strategy is expected to provide a more reliable basis for model comparison than standard Constant Patch Size (CPS) training. To evaluate this assumption, we conducted experiments on the BTCV and KiTS23 datasets, which were identified by~\cite{isensee2024nnu} as the least and most reliable benchmarks for backbone comparison, respectively. Following~\cite{isensee2024nnu}, we compute the ratio between the inter-method and intra-method standard deviations. A higher ratio indicates that architectural differences dominate over random variation, meaning that the dataset and training setup allow for more meaningful model comparison. We used the nnU-Net ResEnc variants (M, L, and XL), which scale both patch size and model capacity, alongside the original nnU-Net configuration. Each model was trained using both CPS and PGPS-Performance sampling strategies, allowing us to assess whether PGPS-Performance improves the signal-to-noise ratio in backbone comparisons.

\textbf{Results:}
Table~\ref{tab:cps_pgps_backbone_comparison} summarizes mean Dice scores across both datasets. For BTCV, the lowest performing nnU-Net variant trained via PGPS-Performance resulted in even higher segmentation performance than all model variants for CPS sampling. Also for KiTS23, PGPS-Performance sampling improved segmentation performance over CPS sampling, while increasing model capacity had a higher impact on the segmentation performance for all variants.

\begin{table}
\centering
\caption{
Comparison of Dice scores between Constant Patch Size (CPS) and Progressive Growing of Patch Size (PGPS) training strategies across nnU-Net and ResEnc variants on BTCV and KiTS23 datasets. Values indicate mean Dice scores over 5-fold cross-validation. Bold values mark the highest performance per architecture and dataset.
}
\label{tab:cps_pgps_backbone_comparison}
\begin{footnotesize}
\begin{tabular}{lcccc}
\toprule
\multirow{2}{*}{\textbf{Model}} & 
\multicolumn{2}{c}{\textbf{BTCV}} & 
\multicolumn{2}{c}{\textbf{KiTS23}} \\
\cmidrule(lr){2-3} \cmidrule(lr){4-5}
 & CPS & PGPS-Perf & CPS & PGPS-Perf \\
\midrule
nnU-Net (orig.) & 83.37 & \textbf{83.81} & 86.02 & \textbf{86.46} \\
ResEncM         & 83.48 & \textbf{83.97} & 87.05 & \textbf{88.00} \\
ResEncL         & 83.30 & \textbf{84.18} & 88.24 & \textbf{88.64} \\
ResEncXL        & 83.30 & \textbf{84.03} & 88.68 & \textbf{88.93} \\
\bottomrule
\end{tabular}
\end{footnotesize}
\end{table}

While PGPS-Performance generally improved absolute performance for all architectures, another critical finding concerns the reduction in statistical variance across training runs and architectures. Table~\ref{tab:inter_intra_variance} shows the intra- and inter-method standard deviations and the resulting inter/intra ratios. On the KiTS23 dataset, previously shown to provide the very stable comparison signal~\cite{isensee2024nnu}, the ratio remains nearly constant (76.4\% vs.\ 76.7\%), indicating that both CPS and PGPS-Performance allow for consistent differentiation of model scales. In contrast, on BTCV, which is known for a low signal-to-noise ratio for backbone comparison, PGPS-Performance substantially increased the ratio from 32.6\% to 55.6\%. This demonstrates that PGPS-Performance effectively increases the signal-to-noise ratio, thereby yielding more reliable and interpretable backbone comparisons. Intra-Method SD is only decreased for KiTS23, while for BTCV it is increased. In our experiment in section~\ref{sec:stochastic-variability}, we focused on repeating training on the same fold. In contrast, here the standard deviation is computed over five folds, such that fold difficulty (annotation errors, sample difficulty) itself also plays an important role.

\begin{table}[b!]
\centering
\caption{
Comparison of Dice score variability across nnU-Net and ResEnc variants for Constant Patch Size (CPS) and Progressive Growing of Patch Size (PGPS) on BTCV and KiTS23. Reported values represent the mean standard deviation of the Dice score over 5-fold cross-validation. The inter/intra-method ratio indicates how clearly the dataset ranks architectures (higher is better). PGPS-Performance increases this ratio, improving backbone comparability.
}
\label{tab:inter_intra_variance}
\renewcommand{\arraystretch}{1.1}
\setlength{\tabcolsep}{5pt}
\begin{footnotesize}
\begin{tabular}{lcccc}
\toprule & 
\multicolumn{2}{c}{\textbf{BTCV}} & 
\multicolumn{2}{c}{\textbf{KiTS23}} \\
\cmidrule(lr){2-3} \cmidrule(lr){4-5}
 & \textbf{CPS} & \textbf{PGPS-Perf} & \textbf{CPS} & \textbf{PGPS-Perf} \\
\midrule
nnU-Net (orig.) & 2.24 & 2.54 & 1.77 & 1.73 \\
ResEncM          & 2.09 & 2.25 & 1.55 & 1.40 \\
ResEncL          & 2.52 & 2.48 & 0.92 & 0.94 \\
ResEncXL         & 2.20 & 2.27 & 1.19 & 0.91 \\
\midrule
Avg. Intra-Method SD [\%] & 2.27 & 2.38 & 1.358 & 1.245 \\
Inter-Method SD [\%] & 0.74 & 1.33 & 1.037 & 0.955 \\
Inter/Intra Ratio [\%] $\uparrow$ & 32.6 & \textbf{55.6} & 76.4 & \textbf{76.7} \\
\bottomrule
\end{tabular}
\end{footnotesize}
\end{table}

Overall, these findings confirm that PGPS-Performance enhances segmentation performance but also improves the robustness of studies comparing different methods or backbones. 

\begin{table*}[h!]
\centering
\caption{
Comparison of Dice scores [\%] across the Medical Segmentation Decathlon (MSD) tasks.
We report results for the baseline constant patch size training (CPS), the \textbf{new PGPS modes} introduced in this work (PGPS-Eff, PGPS-Perf), and the \textbf{original PGPS modes} from MICCAI 2024 (PGPS, PGPS+)~\cite{fischer2024progressive}.
Values are color-coded relative to CPS: \textcolor{blue}{blue} = better than CPS, \textcolor{red}{red} = worse than CPS. \textbf{Bold}: Best performing sampling. \underline{Underlined}: Second best performing sampling. All values are mean Dice scores in 5-fold cross-validation as in~\cite{isensee2019automated}. The newly proposed PGPS-Efficiency is more computationally efficient than the old version, while also yielding a better Dice score. PGPS-Performance achieves the highest Dice score and is the only curriculum outperforming CPS on every task, while it has doubled the computational costs compared to PGPS+. While PGPS-Efficiency is the most efficient version of the four curricula, PGPS-Performance yields the highest segmentation performance.
}
\label{tab:combined_msd_color}
\begin{small}
\begin{tabular}{l ccc ccc}
\toprule
\multirow{2}{*}{\textbf{Dataset}} & \multirow{2}{*}{\textbf{CPS}} 
& \multicolumn{2}{c}{\textbf{New (This Work)}} 
& \multicolumn{2}{c}{\textbf{Old (MICCAI 2024)}} \\
\cmidrule(lr){3-4} \cmidrule(lr){5-6}
 &  & \textbf{PGPS-Eff} & \textbf{PGPS-Perf} & \textbf{PGPS} & \textbf{PGPS+} \\
\midrule
MSD Brain          & 74.12 & \textbf{\textcolor{blue}{74.29}} & \textcolor{blue}{74.15} & \textcolor{blue}{74.12} & \textcolor{blue}{\underline{74.21}} \\
MSD Heart          & \underline{93.29} & \textcolor{red}{93.24} & \textbf{\textcolor{blue}{93.29}} & \textcolor{red}{93.21} & \textcolor{red}{93.28} \\
MSD Liver          & 78.74 & \textcolor{blue}{78.76} & \textbf{\textcolor{blue}{80.60}} & \textcolor{blue}{78.91} & \textcolor{blue}{\underline{79.38}} \\
MSD Hippocampus    & 88.96 & \textcolor{blue}{\underline{89.12}} & \textbf{\textcolor{blue}{89.15}} & \textcolor{blue}{89.11} & \textcolor{blue}{89.07} \\
MSD Prostate       & 73.13 & \textcolor{blue}{75.26} & \textbf{\textcolor{blue}{76.31}} & \textcolor{blue}{\underline{75.66}} & \textcolor{blue}{75.31} \\
MSD Lung           & 70.00 & \textcolor{blue}{72.30} & \textcolor{blue}{\underline{72.77}} & \textcolor{blue}{72.63} & \textbf{\textcolor{blue}{73.33}} \\
MSD Pancreas       & \underline{68.68} & \textcolor{red}{68.60} & \textbf{\textcolor{blue}{68.80}} & \textcolor{red}{68.24} & \textcolor{red}{68.22} \\
MSD Hepatic Vessel & 68.61 & \textcolor{red}{68.05} & \textbf{\textcolor{blue}{68.98}} & \textcolor{red}{67.82} & \textcolor{blue}{\underline{68.71}} \\
MSD Spleen         & \underline{97.02} & \textcolor{red}{95.85} & \textbf{\textcolor{blue}{97.15}} & \textcolor{red}{96.21} & \textcolor{red}{96.54} \\
MSD Colon          & 48.41 & \textcolor{blue}{\underline{50.83}} & \textbf{\textcolor{blue}{51.02}} & \textcolor{blue}{49.25} & \textcolor{blue}{49.67} \\
\midrule
\textbf{Avg. Normalized Dice Score [$\uparrow$]} 
 & 100.00 & \textcolor{blue}{100.94} & \textbf{\textcolor{blue}{101.72}} & \textcolor{blue}{100.66} & \textcolor{blue}{\underline{101.04}} \\
\textbf{Rel. FLOPs [$\downarrow$]} 
 & 100.00 & \textbf{\textcolor{blue}{0.32}} & \textcolor{blue}{0.88} & \textcolor{blue}{\underline{0.35}} & \textcolor{blue}{0.41} \\
\bottomrule
\end{tabular}
\end{small}
\end{table*}


\section{Comparison Between MICCAI 2024 PGPS Modes and Newly Proposed PGPS Modes}
\label{app:changes-pgps}

This section compares the PGPS modes introduced in our MICCAI 2024 work (\textbf{PGPS, PGPS+})~\cite{fischer2024progressive} with the newly proposed variants (\textbf{PGPS-Efficiency, PGPS-Performance}). The aim is to highlight methodological updates and their practical effects.

\subsection{Methodological Differences}

The differences between the MICCAI 2024 and current PGPS modes are:

\begin{itemize}
    \item \textbf{Additional patch size stage:} The new modes disable a PyTorch normalization layer check that previously blocked single-element tensors, enabling an extra patch size stage. This results in slightly higher class balance for both new PGPS curricula and larger batch sizes for PGPS-Performance.
    \item \textbf{Patch size increments:} Instead of fixed sequential axis growth (MICCAI 2024), the new modes always expand the smallest axis first, leading to more balanced spatial scaling and larger average batch sizes.
    \item \textbf{Batch creation strategy:} PGPS+ relies on \textit{Single-Crop}, whereas PGPS-Performance adopts a \textit{Split-Crop} strategy. This enables efficient training with much larger batch sizes by reducing dataloading costs.
    \item \textbf{Batch size policy:} PGPS+ increases batch size such that the input tensor dimensions grow monotonically, while PGPS-Performance fully utilizes GPU memory, increasing the average batch size and class balance.
\end{itemize}

\subsection{Results and Observations}

Table~\ref{tab:combined_msd_color} summarizes Dice scores and relative computational costs for the Medical Segmentation Decathlon tasks. Key observations include:

\begin{itemize}
    \item \textbf{Efficiency:} PGPS-Efficiency reduces relative FLOPs compared to PGPS while achieving slightly higher Dice scores, benefiting from the additional patch size stage and improved class balance.
    \item \textbf{Performance:} PGPS-Performance achieves the highest overall segmentation performance, outperforming CPS on every task and exceeding PGPS+ due to its larger batch size and class balance.
\end{itemize}

Overall, the newly proposed PGPS modes mark a substantial improvement over the MICCAI 2024 versions. PGPS-Efficiency provides the best trade-off between segmentation performance and computational cost. PGPS-Performance delivers the highest segmentation performance while still reducing computational costs compared to CPS. Crucially, the \textit{Split-Crop} strategy in PGPS-Performance drastically reduces dataloading costs, making large-batch 3D segmentation training feasible, whereas \textit{Single-Crop} used in PGPS+ would be prohibitively expensive for large batch sizes in terms of dataloading costs, as seen in Table~\ref{tab:pgps_efficiency} for datasets KiTS23 and AMOS22, yielding training times double as long as for CPS.







\end{document}